\documentclass[runningheads]{llncs}
\usepackage{graphicx}
\usepackage{comment}
\usepackage{amsmath,amssymb} 
\usepackage{color}

\graphicspath{ {./figures/} }
\usepackage{multirow}
\usepackage{makecell}
\usepackage{arydshln}

\newcommand{\splitcell}[1]{\begin{tabular}{@{}l@{}}#1\end{tabular}}
\newcommand{\bsplitcell}[1]{$\left[\splitcell{#1}\right]$}

\usepackage{subfig}
\usepackage{wrapfig}
 \usepackage{floatflt}
\usepackage{hyperref}
\usepackage{xcolor}

\usepackage[width=122mm,left=12mm,paperwidth=146mm,height=193mm,top=12mm,paperheight=217mm]{geometry}

\begin{document}
\pagestyle{headings}
\mainmatter
\def\ECCVSubNumber{1411}  

\title{Improved Residual Networks for \\Image and Video Recognition} 

\titlerunning{Improved Residual Networks}
%
\author{Ionut Cosmin Duta \and
Li Liu \and
Fan Zhu \and
Ling Shao
}
\authorrunning{I.C. Duta et al.}
%
\institute{Inception Institute of Artificial Intelligence (IIAI), Abu Dhabi, UAE\\
\email{\{ionut.duta, li.liu, fan.zhu, ling.shao\}@inceptioniai.org}}
\maketitle

\begin{abstract}
\vspace{-0.1in}
 Residual networks (ResNets) represent a powerful type of convolutional neural network (CNN) architecture, widely adopted and used in various tasks. In this work we propose an improved version of ResNets. Our proposed improvements address all three main components of a ResNet: the flow of information through the network layers, the residual building block, and the projection shortcut. We are able to show consistent improvements in accuracy and learning convergence over the baseline. For instance,  on ImageNet dataset, using the ResNet with 50 layers, for top-1 accuracy  we can report a 1.19\%  improvement over the baseline in one setting and around 2\% boost in another. Importantly, these improvements are obtained without increasing the model complexity. Our proposed approach allows us to train extremely deep networks, while the baseline shows severe optimization issues.
 We report results on three tasks over six datasets: image classification (ImageNet, CIFAR-10 and CIFAR-100), object detection (COCO) and video action recognition (Kinetics-400 and Something-Something-v2). In the deep learning era, we establish a new milestone for the depth of a CNN. We successfully train a 404-layer deep CNN on the ImageNet dataset and a 3002-layer network on CIFAR-10 and CIFAR-100, while the baseline is not able to converge at such extreme depths. Code is available at: {\color{blue}\url{https://github.com/iduta/iresnet}}
 \vspace{-0.02in}
\end{abstract}

\section{Introduction}
\vspace{-0.005in}
We are in the deep learning era. From the beginning of the computer vision revolution \cite{krizhevsky2012imagenet} network depth was highlighted as one of the key factors accountable for obtaining powerful representations that lead to impressive results in numerous tasks.  Over the past several years, the depth of convolutional neural networks (CNNs) \cite{lecun1989backpropagation,lecun1998gradient} has been increasing continuously \cite{krizhevsky2012imagenet,simonyan2014very,szegedy2015going,ioffe2015batch,he2016deep,he2016identity,chollet2017xception,huang2017densely,zoph2018learning}. However, with the increasing depth, optimization/learning difficulties grow as well. Adding more layers does not guarantee better results. Residual networks (ResNets) \cite{he2016deep} exposed the issue of learning very deep CNNs and proposed a solution in terms of residual learning. ResNets are very powerful in learning deep CNNs and widely used in practice, representing the backbone/foundations for various complex tasks, such as object detection and  instance segmentation~\cite{he2016deep,he2017mask,lin2017focal,lin2017feature,xie2017aggregated,hu2018squeeze,wu2018group}. Given the crucial importance of ResNets for learning deep representations for visual recognition, in this paper we investigate new architectures and propose an improved version of ResNets.

ResNets \cite{he2016deep} are composed by stacking together a large number of building blocks.
The core idea of ResNets is to facilitate the learning of the identity mappings for its building blocks, if needed. This is accomplished by using a shortcut/skip connection: adding the input of the block to its learned output. Theoretically, the network can learn  the identity mappings by itself, without these shortcuts. However, in practice, it is not easy for the optimizer to learn identity mappings. This is called the degradation problem. This problem is easily noticeable when training a very deep network, where the accuracy gets worse than for its shallower counterpart, with significantly less layers, even though, theoretically, the situation should be vice-versa, or at least it should not become worse. The ResNet idea is a solution to address the degradation problem, allowing to efficiently learn much deeper networks.
However, the degradation problem has not been completely solved, as pointed out in \cite{he2016identity} and thoroughly verified in our experiments. For instance, increasing the depth from 152 to 200 layers on the ImageNet \cite{russakovsky2015imagenet} dataset leads to significantly worse results, including training error, suggesting severe optimization problems. This shows that ResNet  still harms propagation of information through the network when  the number of layers is increased.  In this work, we propose an improved architecture design that facilitates the  propagation of information through the network. Our design specifically separates the network into stages and applies a different building block depending on the location within each stage. Our proposed architecture is able to learn extremely deep networks, showing no optimization difficulty with increasing depth.

In ResNets, when the dimensions of a building block do not match the dimensions of the next building block, a projection shortcut must be used. The ResNet work \cite{he2016deep} concluded that projection shortcuts are not essential for the degradation problem. However, the projection shortcuts can play an important role in the network architecture, as they are found on the main information propagation path and can thus easily perturb the signal or cause information loss. We introduce an improved shortcut projection, which is parameter-free improvement and shows a significant boost in performance. In the original ResNet~\cite{he2016deep}, a bottleneck building block is introduced to control the number of parameters and computational cost when the depth is considerably increased. However, in this building block architecture, the only convolution responsible for learning spatial filters receives the least number of input/output channels. We propose a building block that changes the focus to the spatial convolution, our architecture contains four times more spatial channels in a building block than the original ResNet~\cite{he2016deep}, while keeping  the number of parameters and computational cost under control.

To summarize, our main contributions are:
({\bf1}) We introduce a network architecture for residual learning based on stages. The proposed approach facilitates the learning process by providing a better path for information to propagate through network's layers (Section \ref{section:stage});
 ({\bf2})~We propose an improved projection shortcut that reduces the information loss and provides better results (see Section \ref{section:proj});
   ({\bf3}) We present a building block that considerably increases the spatial channels for learning more powerful spatial patterns (Section \ref{sec:group});
 ({\bf4})~Our proposed approach provides consistent improvements over the baseline. It is important to note that these improvements are obtained without increasing the model complexity. We present results on six datasets (four for images and two for large-scale video classification). With our network architecture, we are able to effectively train extremely deep CNNs, while the baseline architecture shows  significant optimization issues. We successfully train a 404-layer deep CNN on the ImageNet dataset and a 3002-layer network on CIFAR-10 and CIFAR-100, while the baseline was not able to converge for such depths. To the best of our knowledge, these are the deepest networks ever trained on these datasets (Section \ref{sec:exp}). 
 \vspace{-0.1in}

\section{Related Work}

Residual networks (ResNets) \cite{he2016deep} are very efficient in training deep architectures for visual recognition. ResNets use a shortcut connection to facilitate the signal propagation along the network. We use ResNets as baselines in our work, and show different improved architectures.  There are many works focused on improving this  powerful architecture. The work~\cite{he2016identity} introduces pre-activation ResNets by proposing a new order of the elements for the building block to improve signal propagation along the network. Our work is related to pre-activation ResNets method as one of our contributions addresses also the flow of information through the network. However, different from ~\cite{he2016identity}, our approach specifically splits the network in stages and proposes a different block for each part of the stage: start, middle and end. Our approach shows better results starting from relatively deep network to very deep ones, while ~\cite{he2016identity} shows its benefits mostly on very deep networks. As a matter of fact ~\cite{he2016identity} shows the results starting with the depth 152 on ImageNet dataset~\cite{russakovsky2015imagenet}.

The works~\cite{krizhevsky2012imagenet} uses grouped convolution to split the computation of the convolutions over two GPUs to overcome the limitations in computational resources. The work~\cite{xie2017aggregated} uses also group convolution but with the goal to improve the recognition performance of ResNet architectures.  We make also use of grouped convolution, however, different from ~\cite{xie2017aggregated}, we propose a new building block architecture, with a different shape, which introduces two times more spatial filters than ~\cite{xie2017aggregated}, showing improved performance.
The work \cite{hu2018squeeze} and \cite{wang2018non} improve ResNets by introducing squeeze-and-excitation and non-local blocks. However, different from our work, these are additional blocks that need to be inserted in the network, which increase the model and computational complexity. 
Another work \cite{he2019bag}, focused on improving ResNets,  collects some refinements in a bag of tricks to boost the performance of the network. We propose different directions, and most of their collected refinements do not overlap with our work, they are mostly complementary, therefore, they can be used together for further improvements. Although, there is an overlap in a point with our work, regarding the projection shortcut, our proposed projection shortcut is different than~\cite{he2019bag}. We use a different block, and we include it also in the first stage of the network. In the Appendix  we present a direct comparison of our proposed projection  shortcut with~\cite{he2019bag}.  

\section{Improved residual networks}

\subsection{Improved information flow through the network \label{section:stage}}

\begin{figure}[t]
 \vspace{-0.05in}
  \centering
  \includegraphics[width=0.7\textwidth]{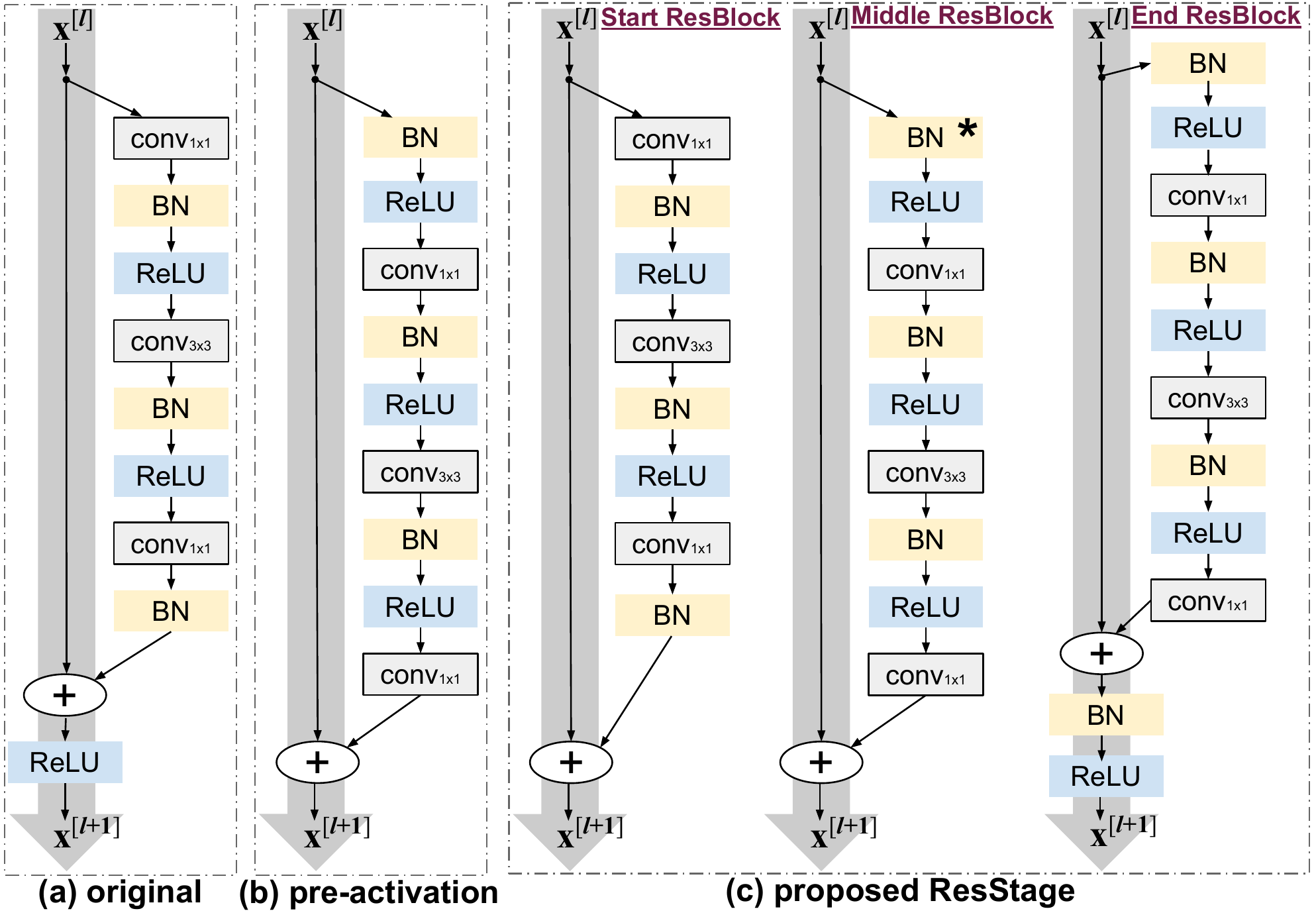}
   \vspace{-0.07in}
  \caption{Residual Building block architectures: (a) original \cite{he2016deep}; (b) pre-activation \cite{he2016identity}; (c) proposed ResStage. ( {\bf$^*$} the first BN in the first Middle Resblock is  eliminated in each stage).}
  \label{fig:resStage}
  \vspace{-0.2in}
\end{figure}

A residual network (ResNet) \cite{he2016deep} is constructed by stacking together many residual building blocks (ResBlocks). An example of a residual block is represented in Fig. \ref{fig:resStage}(a). Each residual block can be formally defined as:
\begin{equation}
\textrm{\bf{x}}^{[l+1]}=  
\left\{\begin{array}{ll}
\mathrm{ReLU}(\mathcal{F}(\textrm{\bf{x}}^{[l]}, \{\mathcal{W}^{[l]}_i \} ) + \textrm{\bf{x}}^{[l]}), & \textrm{if} \;\; \mathrm{size} (\mathcal{F}(\textrm{\bf{x}}^{[l]}, \{\mathcal{W}^{[l]}_i\})) = \mathrm{size}(\textrm{\bf{x}}^{[l]}); \\
\mathrm{ReLU}(\mathcal{F}(\textrm{\bf{x}}^{[l]}, \{\mathcal{W}^{[l]}_i\}) + \mathcal{W}^{[l]}_p\textrm{\bf{x}}^{[l]}), & \textrm{if} \;\; \mathrm{size} (\mathcal{F}(\textrm{\bf{x}}^{[l]}, \{\mathcal{W}^{[l]}_i\})) \neq \mathrm{size}(\textrm{\bf{x}}^{[l]}),
\end{array}\right.
\label{eq:res}
\end{equation}
where $\textrm{\bf{x}}^{[l]}$ and $\textrm{\bf{x}}^{[l+1]}$ are the input and output vectors of the $l$-th ResBlock; $ReLU$ represents the activation function \cite{nair2010rectified};  $\mathcal{F}(\textrm{\bf{x}}^{[l]}, \{\mathcal{W}^{[l]}_i \} )$ is a learnable residual mapping function that can have several layers (indexed be $i$), for instance, for a bottleneck ResBlock $l$ with three layers, as in~\cite{he2016deep}, $\mathcal{F} = \mathcal{W}^{[l]}_3 \mathrm{ReLU}(\mathcal{W}^{[l]}_2 \mathrm{ReLU} (\mathcal{W}^{[l]}_1\textrm{\bf{x}}))$;  $\mathcal{W}^{[l]}_p$ is a learnable linear  projection matrix that allows mapping the size of $\textrm{\bf{x}}^{[l]}$  to the output size of  $\mathcal{F}$, which exists only in case of not corresponding dimensions for performing the element-wise addition between  $\mathcal{F}$ and $\textrm{\bf{x}}^{[l]}$.

ResNets are specifically created to easily allow information to propagate forward and backward through the network. The original ResBlock bottleneck  of \cite{he2016deep}, illustrated in Fig.~\ref{fig:resStage}(a), consists of three convolutional layers (two with a kernel $1\times1$ and one with a $3\times3$), three batch normalizations~\cite{ioffe2015batch} and  three ReLU layers. 
In the original ResBlock, the big grey arrow represents the most direct path for information to propagate (which includes the shortcut connection). However, as we can see in Fig.~\ref{fig:resStage}(a) and in Equation~(\ref{eq:res}), there is a ReLU activation function on the main propagation path. This ReLU can potentially  negatively affect the propagation of information by zeroing the negative signal. This is especially critical at the beginning of the training (after a while the network may start adapting the weights to output a positive signal that is not affected when passing through a ReLU). This aspect is investigated in \cite{he2016identity} (see Fig.~\ref{fig:resStage}(b)) where they proposed a redesigned ResBlock, called pre-activation, by moving the last BN layer and ReLU to the beginning.

 On one extreme we have original ResNet~\cite{he2016deep} (which hampers the signal propagation with too many gates (e.g. ReLUs) on the main path), on the other extreme, there is pre-act. ResNet~\cite{he2016identity} (which allows the signal to pass through the network in an uncontrolled way). Both extremes are not optimal, and present different issues. Leaving the main path completely free (as in pre-act. ResNet~\cite{he2016identity}) raises two main issues. First, note that over all four stages, there is no normalization (BN) of the full signal (all BN are applied  independently on the branches, but not on the full signal, after addition), thus, as we add more blocks the full signal becomes more "unnormalized", this creates difficulties in learning. This issue is present for both original ResNet~\cite{he2016deep} and pre-act. \cite{he2016identity}.
 Second, note that there are four projections shortcuts (thus four 1x1 conv on the main path), theoretically, according to \cite{he2016identity}, the network can learn the identity mappings for most of the blocks  (the branch in the building block with convolutions can output zeros, to easily create the identity when we apply the addition operation,  this is actually a common initialization \cite{goyal2017accurate}). In this case, the pre-act. ResNet over all four main stages ends-up with only four successive 1x1 conv (from the projection shortcut in the main path) but without any non-linearity in between, limiting the learning capability.  Our approach addresses also these two issues,  as it stabilizes the signal before each main stage (we use a BN on the full signal after each main stage) and ensures that there is at least one non-linearity (applied on the full signal) at the end of each stage.

Our proposed ResBlock is illustrated in Fig. \ref{fig:resStage}(c).
The ResNet can be split into different stages.  As a concrete example, we can take the ResNet with a depth of 50 (ResNet-50) illustrated in Table \ref{table:net}; however, this can be extended to any depth. A possible separation into stages for ResNet-50 is determined by the output spatial size and the number of output channels. When either the output spatial size or number of output channels is going to change, it marks the start of another stage. For the ResNet-50, we obtain four main stages (which contain ResBlocks) and a starting and ending stage. Each of the four main stages can contain a number of ResBlocks; in the case of ResNet-50, there are three ResBlocks for stage 1, four for stage 2, six for stage 3 and three for stage 4. Each main stage is divided into three parts: one Start ResBlock, a number of Middle ResBlocks (which can be any number; in the case of ResNet-50 there are [1, 2, 4, 1] Middle ResBlocks for the corresponding stages) and one End ResBlock. Each ResBlock has a different design depending on the position in the stage.   We call ResStage network the results of splitting a ResNet into the proposed stages architecture.

It is important to point out that our proposed solution does not increase the model complexity. For instance, on the ResNet-50, all three approaches (the original \cite{he2016deep}, pre-activation \cite{he2016identity} and our proposed ResStage) contain the same number of components and the same number of parameters. Only the arrangement of the components is changed. Note that, for our proposed ResStage, in each main stage, we eliminate the first BN in the first Middle ResBlock, as the signal is already normalized by our Start ResBlock.

Different from the original approach \cite{he2016deep}, our proposed ResStage contains a fixed number of ReLUs on the main path for the information to propagate forward and backward. For instance, the number of ReLUs on the main propagation path  in \cite{he2016deep} is directly proportional to the network depth. While in our ResStage, for the main stages,  there are only four ReLUs on the main information propagation path, which are not influenced by changing the depth. This enables the network to avoid hampering the signal when information is passed through numerous layers.

Different from \cite{he2016identity}, we specifically split the networks into several stages, each of which contains three parts. The End Resblock of each stage is completed with a BN and ReLU, which can be seen as preparation for the next stage, stabilizing and preparing the signal to enter into a new stage. In our Start ResBlock, there is a BN layer after the last conv, which normalizes the signal, preparing it for the element-wise addition with the projection shortcut (which also provides a normalized signal). 
Our proposed approach facilitates  learning by   offering a better path for information to propagate through the network. ResStage eases optimization, allowing extremely deep networks to be easily trained. The network can easily and dynamically choose which ResBlocks to use and which to discard (by easily setting the weights towards  0) during the learning process.

 Our proposal for learning in stages (with 3 main parts for each stage, similar to the parts of a story: introduction, body and conclusion) is  designed for efficient information flow but also for keeping the signal under control.
In the experimental section we show the gain in performance of our model over the baseline \cite{he2016deep} and \cite{he2016identity}, while maintaining a similar model complexity.

\subsection{Improved projection shortcut\label{section:proj}}
\begin{wrapfigure}{r}{0.19\textwidth}
\vspace{-0.3in}
  \centering
    \includegraphics[width=0.20\textwidth]{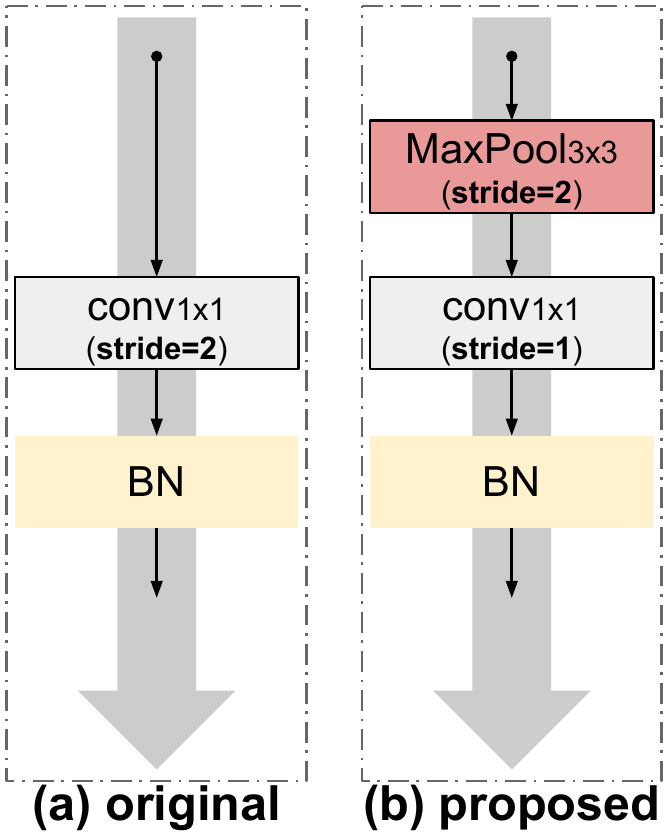}
  \caption{(a) original projection shortcut \cite{he2016deep}; (b) proposed projection shortcut.}
  \label{fig:proj}
  \vspace{-0.3in}
\end{wrapfigure}

In the original ResNet architecture \cite{he2016deep}, when the dimensions  of $\textrm{\bf{x}}$ do not match the output dimensions of  $\mathcal{F}$, then a projection shortcut is applied to $\textrm{\bf{x}}$  (instead of an  identity shortcut, see Equation \ref{eq:res}) to make the element-wise addition possible.
The default projection shortcut used in the original ResNet~\cite{he2016deep} is illustrated in Fig. \ref{fig:proj}(a).\footnote{ Note that \cite{he2016deep} introduced three projection shortcut options: A, B, C. However, B is widely used and adopted as the default option, as option A negatively affects the accuracy and option C  considerably increases the model complexity (even doubled).}
The original projection shortcut uses a conv with a 1$\times$1 kernel to project the channels of $\textrm{\bf{x}}$ to the number of output channels of $\mathcal{F}$. Note that the stride of the 1$\times$1 conv is two, aligning the spatial sizes between $\textrm{\bf{x}}$ and the output of $\mathcal{F}$. Then, a BN is applied before the element-wise addition with the  output of $\mathcal{F}$.  Therefore, both channel and spatial matching are performed by a 1$\times$1 conv. This causes a significant loss in information as the 1$\times$1 conv (with stride~2) skips 75\% of the feature maps activations when reducing the spatial size by two. Furthermore, there is no meaningful criterion for selecting the 25\% of the feature maps activations considered by the 1$\times$1 conv.
The result is then  added to the  main ResBlock output. Thus, this noisy output of the projection shortcut contributes with relatively half of the  information to the next ResBlock. This introduces noise and information loss, and can negatively perturb the main flow of the information through the network. 

Our proposed projection shortcut is presented in Fig. \ref{fig:proj}(b). We disentangle the spatial projection from the channel projection. For the spatial projection, we perform a 3$\times$3 max pooling with stride 2. Then, we apply a 1$\times$1 conv with stride 1 for the channel projection, followed by a BN. With the max pooling, we introduce a criterion for selecting which activations should be considered for 1$\times$1 conv. Furthermore, the spatial projection considers all the information from the feature maps and picks the element with the highest activation to be considered in the next step.  Note that the kernel of the max pooling coincides with the kernel of the middle conv from the ResBlock, which ensures that the element-wise addition is performed between elements computed over the same spatial window. Our proposed projection shortcut reduces the information loss and, in the experimental section, we show the benefit in performance.

Besides the first motivation to reduce the information loss and the perturbation of the signal, there are other two reasons for our proposed projection shortcut. Second, having a max pooling on Start ResBlock of each main stage improves the translation invariance of the network and ultimately improves the overall recognition performance. Third motivation: with our projection shortcut, the Start ResBlock of each stage, which performs the downsampling, can be seen as a combination between "soft downsampling" (or weighted downsampling), accomplished by the 3x3 conv (with learnable weights for downsampling) of the Start ResBlock,  and "hard downsampling", accomplished by the 3x3 max pooling of our projection shortcut. For illustration, we can force to make a parallel view between soft assignment (see in Fisher Vectors encoding~\cite{perronnin2010improving}) and hard assignment (see in VLAD encoding~\cite{jegou2011aggregating}). Each of the two downsamplings comes with complementary benefits. The "hard downsampling" is beneficial for classification (picking the element with the highest activation), while the "soft downsampling" is also a contribution to not losing all spacial context (therefore, helping for better localization, as the transition between elements is smoother).

Note that our proposed projection shortcut does not add any additional parameters to the model. A ResNet  usually requires only four projection shortcuts, at the beginning of each stage (on the Start ResBlock of a stage). Therefore, for our proposed projection shortcut the increase in computational cost for ResNet is negligible, since we need to additionally include only three max pooling layers (as for the first stage we use the existing max pooling in the ResNet), which are usually cheap to compute. We refer to the improved residual network ({\bf \mbox{iResNet}}) when using   the idea of stages from the previous section together with the proposed projection shortcut.

\subsection{Grouped building block \label{sec:group}}
The bottleneck building block was introduced in \cite{he2016deep} for practical considerations, to maintain a reasonable computational cost when increasing the network depth. It first contains a 1$\times$1 conv for reducing the number of channels, then a 3$\times$3 conv bottleneck which operates on  the smallest number of input/output channels, and, finally, a 1$\times$1 conv that increases  the number of channels back to the original. The reason for this design is to run the 3$\times$3 conv on the smaller number of channels to keep  computational cost and the number of parameters under control. However, the 3$\times$3 conv is very important as it is the only component able to learn spatial patterns, but in the bottleneck design it receives the smaller number of input/output channels. 

We propose an improved building block that contains the biggest number of input/output channels on the 3$\times$3 conv. In the design of our proposed building block we make use of grouped convolution, we call it ResGroup block. Grouped convolution was used in \cite{krizhevsky2012imagenet} as a solution to distribute the model over two GPUs to overcome the limitations caused by computational cost and memory. Recently, \cite{xie2017aggregated} exploited grouped convolution to improve accuracy. For the standard convolution, each output channel is connected to all input channels. The main idea of grouped convolution is to split the input channels into several groups and perform convolution operations independently for each group. In this way, the number of parameters (params) and floating-point operations (FLOPs) can be reduced by a factor equal to the number of groups. The number of params and FLOPs (on image data) can be computed as:
\begin{equation}
    params = \frac{ch_{in}}{G} \cdot ch_{out} \cdot  k_1 \cdot k_2; \;\;\;\: FLOPs = \frac{ch_{in}}{G} \cdot ch_{out} \cdot  k_1 \cdot k_2 \cdot w \cdot h,
\end{equation}
where, $ch_{in}$ and $ch_{out}$ are the number of input and output channels; $k_1$ and $k_2$ represent the kernel size of the conv; $w$ and $h$ are the width and height of the channels; and $G$ represents the number of groups the channels are split into. If $G=1$, then we have a standard convolution. If $G$ is equal to the number of input channels, then we are at the other extreme, which is called depthwise convolution, used successfully in \cite{chollet2017xception}.

\begin{table}[t]
\centering
\caption{ Proposed ResGroup and ResGroupFix architectures.}
\label{table:net}
\scalebox{0.8}{
\begin{tabular}{c|c|c|c|c}
\hline
stage                  & output                 & ResNet-50             & ResGroupFix-50              & ResGroup-50              \\ \hline
\multirow{2}{*}{starting}                 & 112$\times$112                & 7$\times$7, 64, stride 2     & 7$\times$7, 64, stride 2     & 7$\times$7, 64, stride 2   \\  \cline{2-5}
 & 56$\times$56 & 3$\times$3 max pool, stride2 & 3$\times$3 max pool, stride2 & 3$\times$3 max pool, stride2 \\  \hline
1&56$\times$56 & \bsplitcell{1$\times$1, 64\\ 3$\times$3, 64\\ 1$\times$1, 256}$\times$3 &\bsplitcell{1$\times$1, 256\\ 3$\times$3, 256, G=64\\ 1$\times$1, 128}$\times$3 & \bsplitcell{1$\times$1, 256\\ 3$\times$3, 256, G=8\\ 1$\times$1, 128}$\times$3 \\ \hline

2 & 28$\times$28 & \bsplitcell{1$\times$1, 128\\ 3$\times$3, 128\\ 1$\times$1, 512}$\times$4 &\bsplitcell{1$\times$1, 512\\ 3$\times$3, 512, G=64\\ 1$\times$1, 256}$\times$4 & \bsplitcell{1$\times$1, 512\\ 3$\times$3, 512, G=16\\ 1$\times$1, 256}$\times$4 \\ \hline

3 & 14$\times$14 & \bsplitcell{1$\times$1, 256\\ 3$\times$3, 256\\ 1$\times$1, 1024}$\times$6 &\bsplitcell{1$\times$1, 1024\\ 3$\times$3, 1024, G=64\\ 1$\times$1, 512}$\times$6 & \bsplitcell{1$\times$1, 1024\\ 3$\times$3, 1024, G=32\\ 1$\times$1, 512}$\times$6 \\ \hline

4 & 7$\times$7 & \bsplitcell{1$\times$1, 512\\ 3$\times$3, 512\\ 1$\times$1, 2048}$\times$3 &\bsplitcell{1$\times$1, 2048\\ 3$\times$3, 2048, G=64\\ 1$\times$1, 1024}$\times$3 & \bsplitcell{1$\times$1, 2048\\ 3$\times$3, 2048, G=64\\ 1$\times$1, 1024}$\times$3 \\ \hline
ending & 1$\times$1 &  \begin{tabular}[c]{@{}c@{}}global avg pool\\ 1000-d fc\end{tabular} & \begin{tabular}[c]{@{}c@{}}global avg pool\\ 1000-d fc\end{tabular} &\begin{tabular}[c]{@{}c@{}}global avg pool\\ 1000-d fc\end{tabular} \\ \hline
\multicolumn{2}{c|}{\# params}                  &    \bf{25.56}  $\times$ $10^6$                 &   \bf{23.37}  $\times$ $10^6$                    &    \bf{24.89}  $\times$ $10^6$                   \\ \hline
\multicolumn{2}{c|}{FLOPs}                      &    \bf{4.14}  $\times$ $10^9$                   &  \bf{4.30}  $\times$ $10^9$&          \bf{5.43}  $\times$ $10^9$             \\ \hline
\end{tabular}}
\end{table}

The proposed network architecture, which has a  similar computational cost and number of parameters as the original ResNet-50 \cite{he2016deep}, is illustrated in Table~\ref{table:net}. To keep  the number of parameters and computational cost under control, the grouped convolution is used with the 3$\times$3 spatial kernel.
Note that each convolution involves a BN and a ReLU (not represented for simplicity). We propose two architectures: (1) ResGroupFix-50 represents the case where the number of groups for each stage is fixed (64 in our case). This option generates a similar number of FLOPs and 8.57\% less  parameters than the baseline \cite{he2016deep} for 50 layers; (2)~ResGroup-50 represents the case where we adapt the number of groups to the channels,  in such a way that all stages have the same number of channels per group (32 in our case). ResGroup-50 has a similar number of parameters as the original ResNet-50 and more FLOPs. 

\begin{wrapfigure}{r}{0.28\textwidth}
  \centering
    \includegraphics[width=0.25\textwidth]{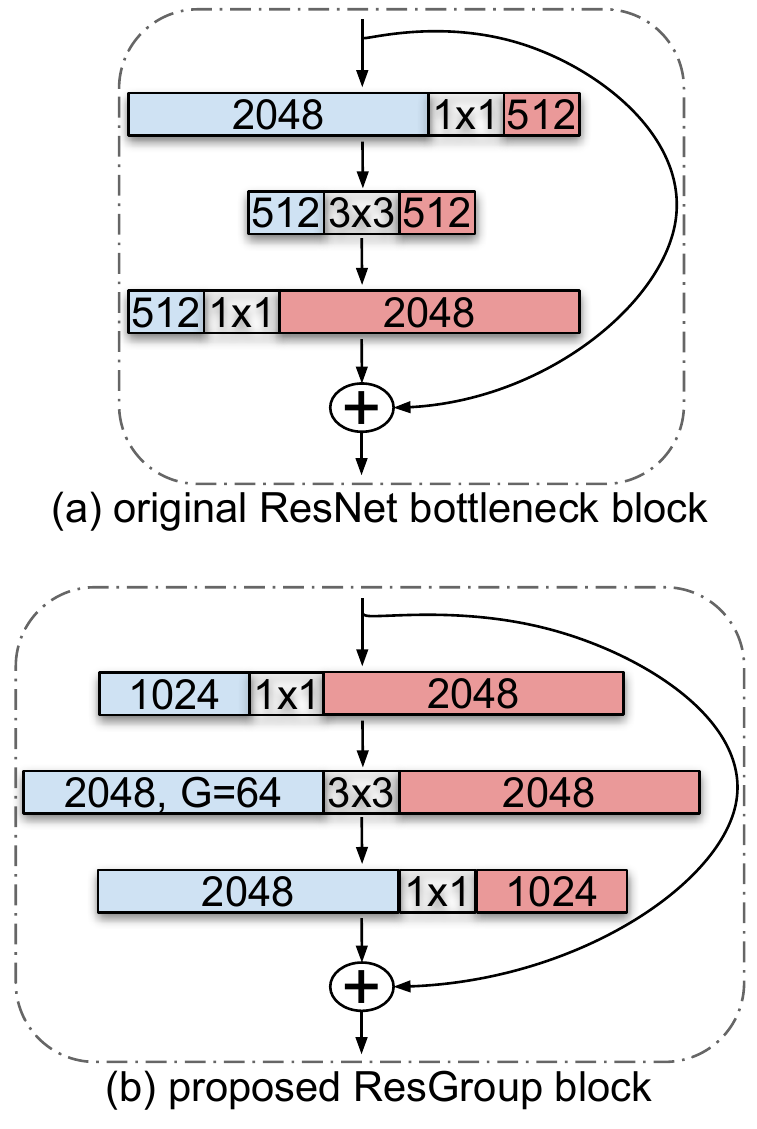}
\captionof{figure}{\small ResGroup block.\label{fig:res_group_block}}
\end{wrapfigure}
Different from the original ResNet \cite{he2016deep} and from ResNeXt \cite{xie2017aggregated} which  use a bottleneck building block, in our case we use a ResGroup block, which changes the focus from the  1$\times$1 conv to the 3$\times$3, see Fig.~\ref{fig:res_group_block} for illustrative comparison on the last residual building block of the network. With this approach, the 3$\times$3 has  the biggest number of channels and a higher ability to learn spatial patterns. Our approach introduces four times more spatial channels than the original \cite{he2016deep} and two times more than~\cite{xie2017aggregated}.  

Both, ResNet~\cite{he2016deep} and ResNeXt~\cite{he2016identity}, use a bottleneck shape for the building block where the spatial conv (3x3) runs on the smallest number of channels (1x1 conv has the largest number of channels), while in our proposed block shape, the  3$\times$3 conv runs on the largest number of channels (see Fig.~\ref{fig:res_group_block}). This is important for improving the performance as the  3$\times$3 conv is the only component responsible for learning spatial patterns, thus, giving the largest number of channels to the  3$\times$3 conv  improves  the capabilities for detecting spatial patterns.
In the experimental section we show the gain in accuracy of our approach over the baseline \cite{he2016deep} and ResNeXt~\cite{xie2017aggregated}.

\section{Experiments \label{sec:exp}}

{\bf Experimental setup.} We report results over six datasets. ImageNet \cite{russakovsky2015imagenet} is the main dataset used in our experiments. It consists of 1000 classes, 1.28 million training images and 50k  validation images. We report both top-1 and top-5 error rates. The CIFAR-10 \cite{krizhevsky2009learning} dataset contains 50k training images and 10k testing images, with 10 classes. CIFAR-100 \cite{krizhevsky2009learning} contains 50k training images and 10k testing images, with 100 classes. On these two datasets we report top-1 error. Kinetics-400 \cite{kay2017kinetics} is a large-scale video recognition dataset that contains  $\sim$246k training videos and 20k validation videos, with 400 action classes. Something-Something-v2 \cite{goyal2017something,mahdisoltani2018effectiveness} is an action recognition dataset focused on modeling temporal relationships. It consists of 168,913 training videos  and 24,777 validation videos. On both video datasets we report top-1 and top-5 error rates. For all these datasets, we follow the standard training and testing protocols. We follow the settings in \cite{goyal2017accurate,he2016deep,he2016identity} and, on all datasets, use the SGD optimizer with a standard momentum of 0.9 and weight decay of 0.0001. For ImageNet and on both video datasets, we train the model for 90 epochs, starting with a learning rate of 0.1 and reducing it by 1/10 at the 30-th, 60-th and 80-th epochs, similar to \cite{he2016deep,goyal2017accurate}; the models are trained over 8 GPUs V100.  For ImageNet, we use the standard 256 training mini-batch size and data augmentation as in~\cite{szegedy2015going,goyal2017accurate}. For videos, we use a minibatch of 64 clips. Data augmentation for videos is similar to~\cite{simonyan2014very,wang2018non} and for CIFAR-10/100 as \cite{he2016deep}.  Refer to the Appendix for more details.
On object detection task we use COCO dataset~\cite{lin2014microsoft} which contains 80 object categories. The models are trained on COCO train2017 (118K images) and tested on val2017 (5K images).  We use a SGD optimizer with momentum 0.9, weight decay 0.0005. We train for 130 epochs using 8 GPUs with 32 batch size each (overall 60K training iterations). We start with a leaning rate of 0.02 and reduce it by 1/10 before 86-th and 108-th epochs. We use a linear warmup during the first epoch ~\cite{goyal2017accurate}. As data augmentation, we perform random crop as in \cite{liu2016ssd}, random horizontal flip and color jitter. We report the metrics exactly as in~\cite{liu2016ssd} using the input image size of $300$$\times$$300$. We use PyTorch library~\cite{paszke2019pytorch}.

\addtolength{\tabcolsep}{+1pt}
\begin{table}[t]
\centering
\caption{Validation error rates (\%) comparison results of iResNet on ImageNet.}
\label{table:res_224}
\begin{tabular}{l|cccc|cccc}
\hline
\multirow{2}{*}{Network}&\multicolumn{4}{c|}{50 layers}&\multicolumn{4}{c}{ 101 layers}\\ \cline{2-9}
&top-1&top-5&params&\small GFLOPs&top-1&top-5&params&\small GFLOPs\\ \hline
baseline \cite{he2016deep}&23.88&7.06&25.56&4.14&22.00&6.10&44.55&7.88\\ 
pre-activation \cite{he2016identity}&23.77&7.04&25.56&4.14&22.11&6.26&44.55&7.88\\ 
ResStage&23.25&6.81&25.56&4.14&21.75&6.01&44.55&7.88\\ 
iResNet&\bf22.69&\bf6.46&25.56&4.18&\bf21.36&\bf5.63&44.55&7.92\\ 
\hline
\hline
&\multicolumn{4}{c|}{ 152 layers}&\multicolumn{4}{c}{ 200 layers}\\ \cline{2-9}
&top-1&top-5&params&\small GFLOPs&top-1&top-5&params&\small GFLOPs\\ \hline
baseline \cite{he2016deep}&21.55&5.74&60.19&11.62&22.45&6.39&64.67&15.16\\ 
pre-activation \cite{he2016identity}&21.41&5.78&60.19&11.62&21.29&5.67&64.67&15.16\\ 
ResStage&21.03&5.65&60.19&11.62&20.88&5.57&64.67&15.16\\  
iResNet&\bf20.66&\bf5.43&60.19&11.65&\bf20.52&\bf5.36&64.67&15.19\\  \hline
\end{tabular}
\end{table}
\addtolength{\tabcolsep}{-1pt}

{\bf Results of iResNet on ImageNet.} Table \ref{table:res_224} shows comparison results  with the baseline ResNet~\cite{he2016deep} and \cite{he2016identity}  over 50-, 101-, 152- and 200-layer deep networks. In all configurations, our proposed ResStage  outperforms the baseline \cite{he2016deep} and~\cite{he2016identity}. Adding our improved shortcut projection to ResStage forms iResNet, which further improves  the results, outperforming significantly \cite{he2016deep} and \cite{he2016identity}. For instance, for 50 layers, iResNet reduces the top-1 error by 1.19\% compared to baseline  \cite{he2016deep} and by 1.08\% in comparison to~\cite{he2016identity}. On the 200-layer network, the degradation problem begins to be visible for the baseline. Increasing the depth from 152 to 200 layers provides better results for our approach. However, for the baseline, the validation results, and also the training results, are considerably worse. For baseline the results for 200 layers are even worse than the setting with 101 layers. This shows severe optimization issues, suggesting that the ResNet architecture harms the signal when it is propagated through many layers. In contrast, iResNet continuously improves  the results with increasing depth. On the 200-layer network, iResNet outperforms ResNet by a large margin, by 1.93\% top-1 error.

\addtolength{\tabcolsep}{-3pt}
\begin{figure*}[t]
\centering
\begin{tabular}{cccc}
\subfloat{\includegraphics[width=0.26\textwidth]{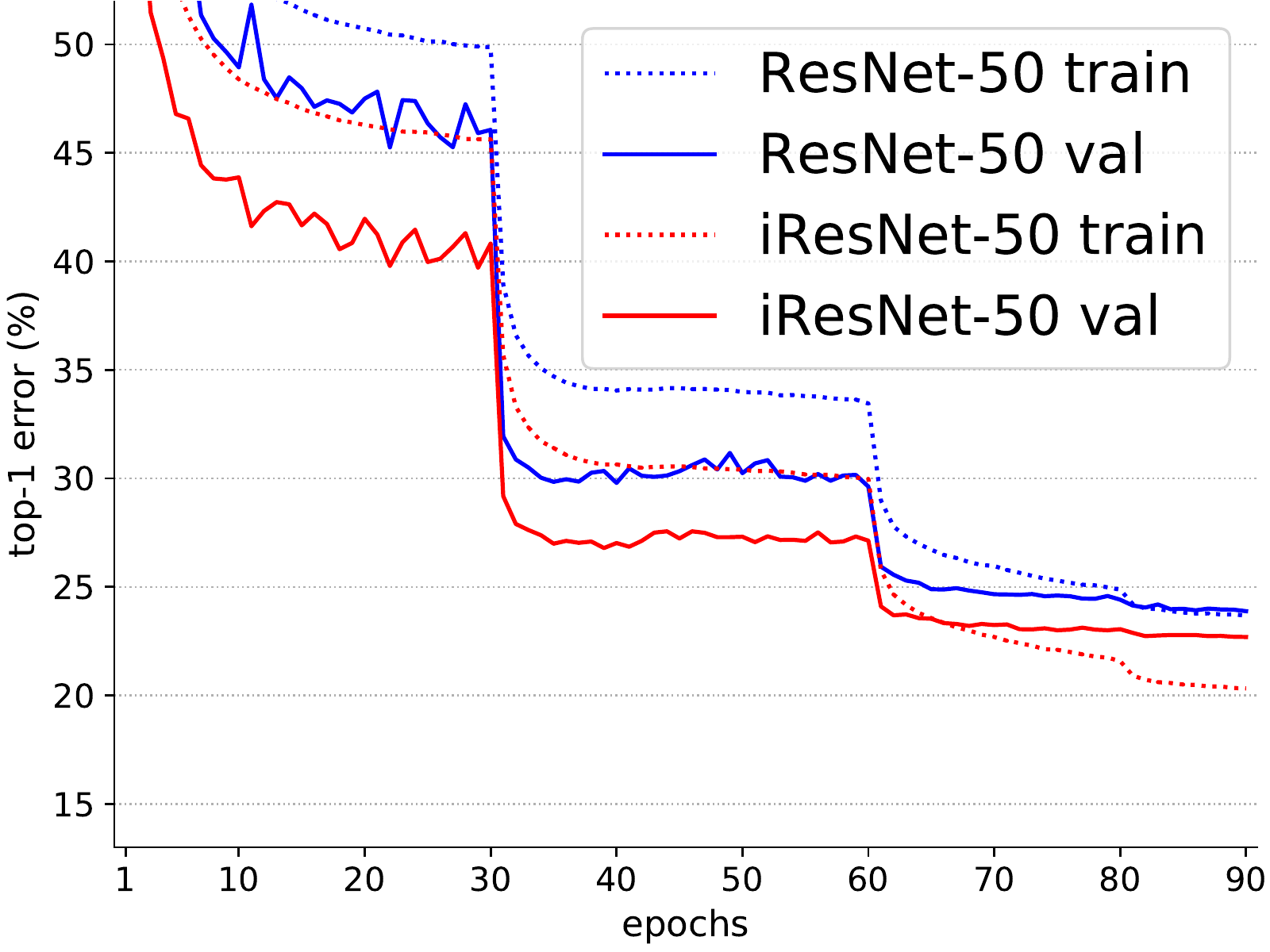}} &
\subfloat{\includegraphics[width=0.26\textwidth]{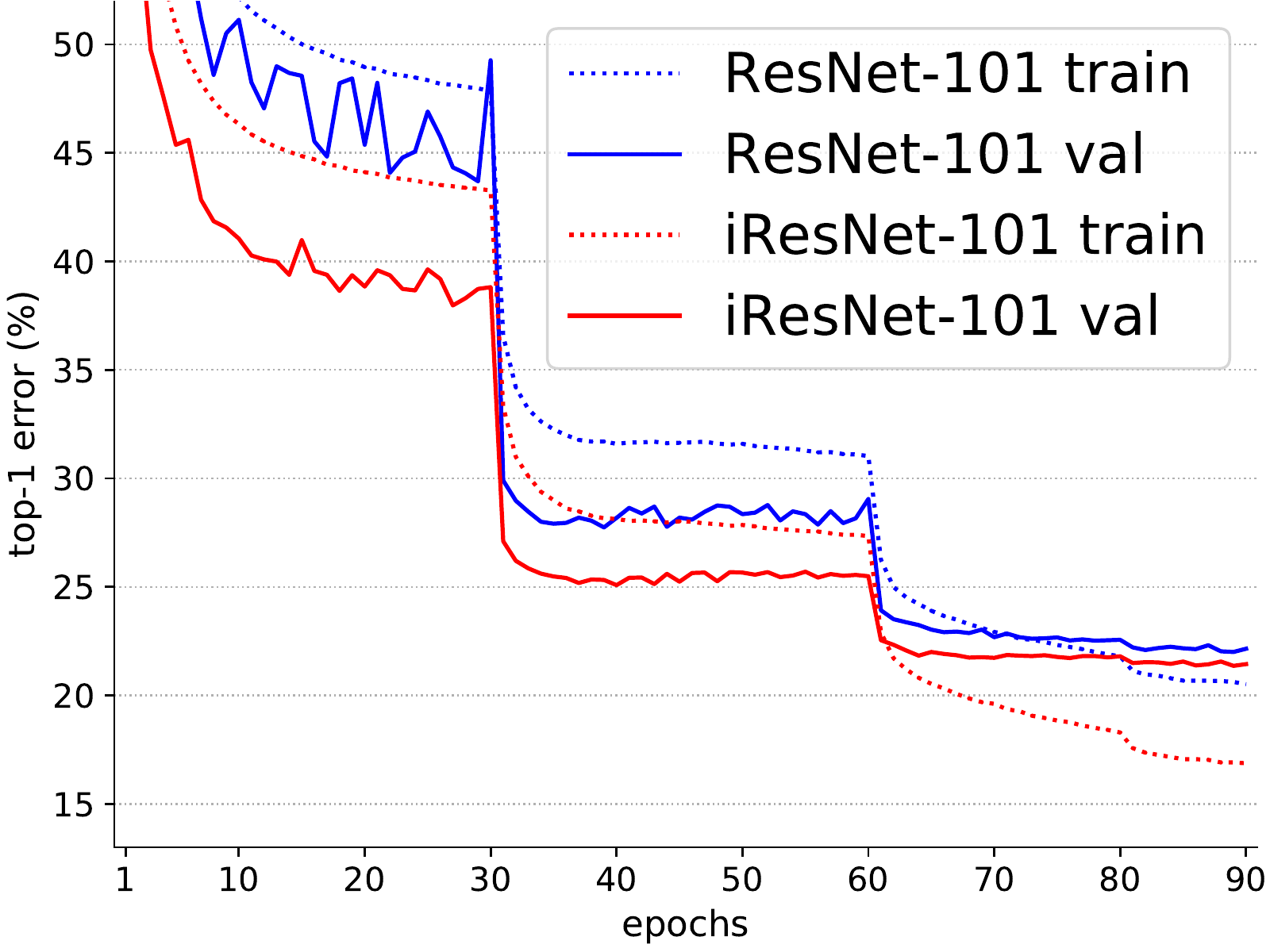}} &
\subfloat{\includegraphics[width=0.26\textwidth]{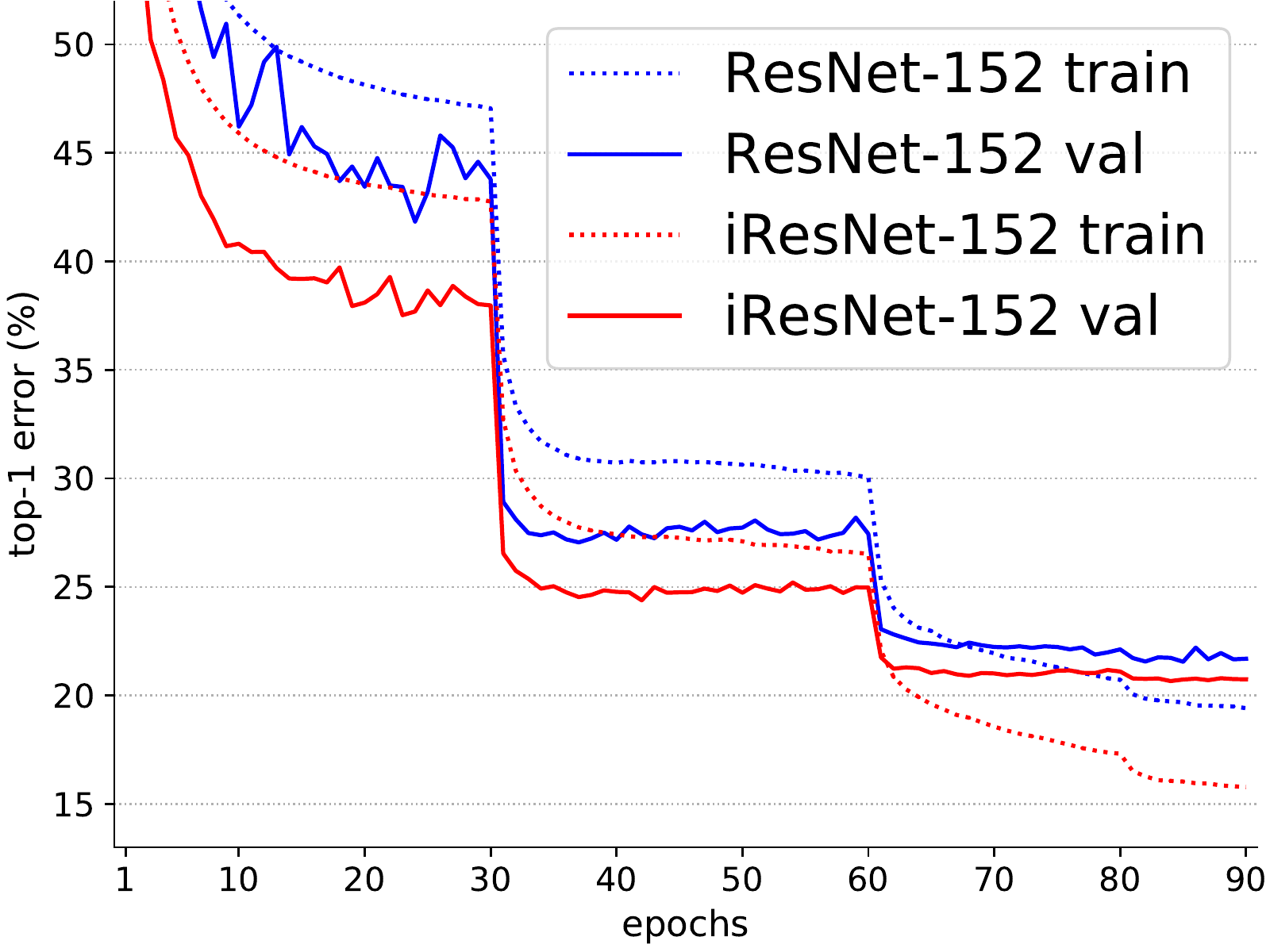}} &
\subfloat{\includegraphics[width=0.26\textwidth]{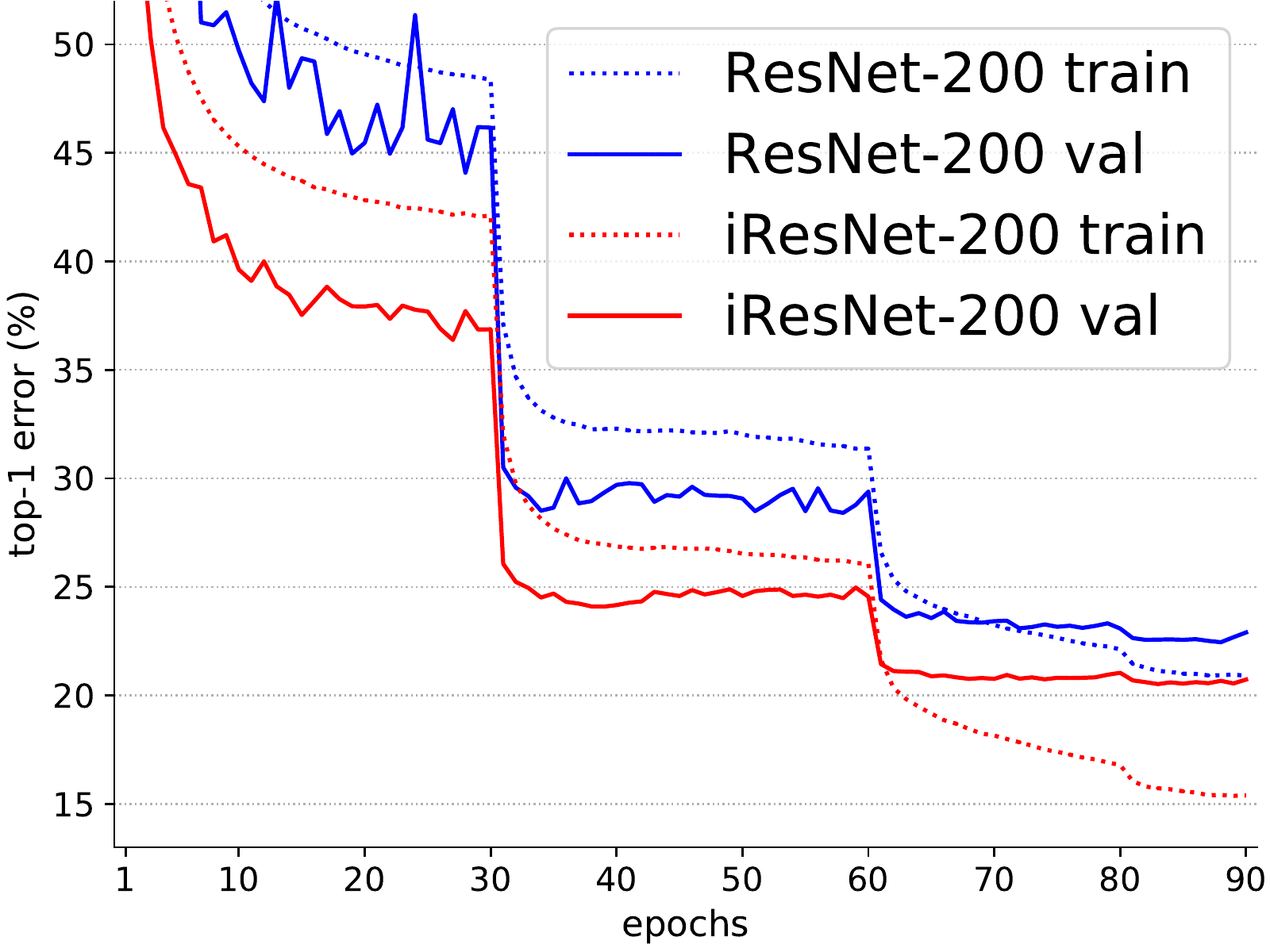}}
\end{tabular}
\caption{Training and validation curves on ImageNet for ResNet and iResNet on 50, 101, 152 and 200 layers.}
\label{fig:curves_iresnet}
\end{figure*}
\addtolength{\tabcolsep}{3pt}

Fig. \ref{fig:curves_iresnet} shows the learning curves comparing our iResNet with the baseline for 50-, 101-, 152- and 200-layer deep networks. The training error decreases with increasing depth for our iResNet, showing the success of optimization. 
The proposed  iResNet speeds-up the learning process. For instance, on 50 layers, on the first interval (first 30 epochs, before the first reduction of learning rate), our iResNet needs less than 8 epochs to outperform the best results of ResNet~\cite{he2016deep} on all first 30 epochs. When increasing the depth the difference is even more severe. Thus, our approach improves training convergence and can require significantly less epochs for training to outperform the baseline. 
Importantly, all these improvements of iResNet are obtained nearly for free, without increasing model complexity.
As our approach does not show any optimization issues on these depths, we ask: What are the limits for iResNet in terms of depth, at what depth will the degradation issue present itself, resulting in an increased training error? To answer this question, we increase the number of layers to 302. The results are presented in Table~\ref{table:extreme_depth}. We also try to train the baseline ResNet with this depth, but the network is not able to converge to a good result, showing severe optimization problems. When increasing to 302 layers, our network still shows no optimization issues, as the increase in depth yields  improved performance on training and validation error. We take a step even further and increase the depth to 404 layers.  See Fig. \ref{fig:curves_extreme_depth} for training curves while increasing depth. Even with this extreme depth, our iResNet provides improved results, showing still no convergence issue. Therefore, we cannot find the depth limits for our approach. The only limits that we do find are in terms of computational resources and time.

\begin{table*}[t]
\parbox{0.39\textwidth}{
\vspace{-0.3in}
\centering
\caption{{\small Error rates results of iResNet on ImageNet with extreme depth: 302 and 404 layers.  P stands for parameters}.}
\label{table:extreme_depth}
\scalebox{0.9}{
\begin{tabular}{l|ccc}
\hline
Network&top-1&top-5&\scalebox{0.7}{ P/GFLOPs}\\ \hline
iResNet-302 & 20.45& 5.28 &96.59/22.67 \\
iResNet-404& \bf20.30&  \bf5.26&  124.5/30.15 \\
\hline
\end{tabular}}
}
\parbox{0.29\textwidth}{
\centering
\includegraphics[width=0.29\textwidth]{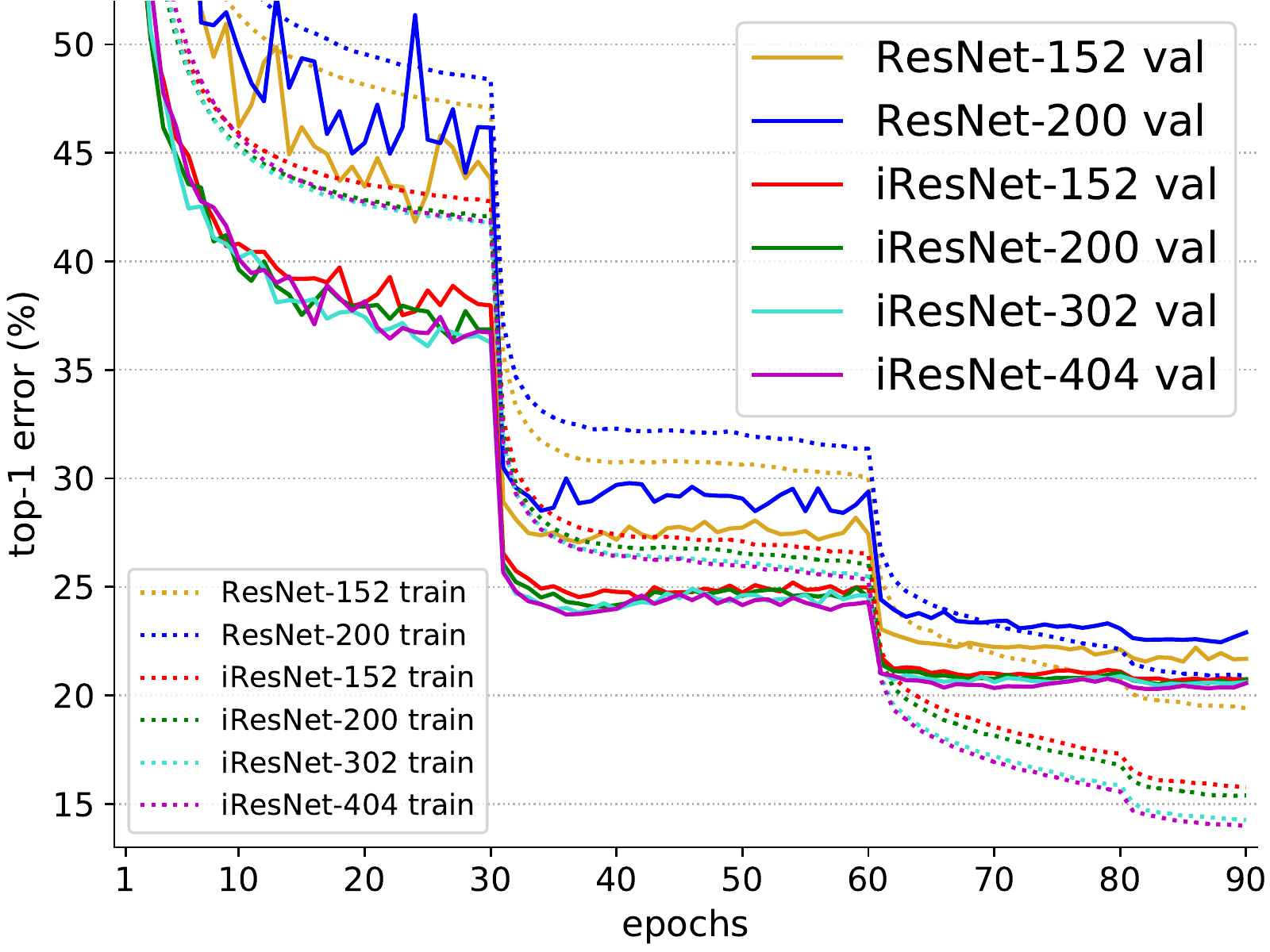}
\captionof{figure}{\small Extreme depths.\label{fig:curves_extreme_depth}}
}
\parbox{0.29\textwidth}{
\centering
\includegraphics[width=0.29\textwidth]{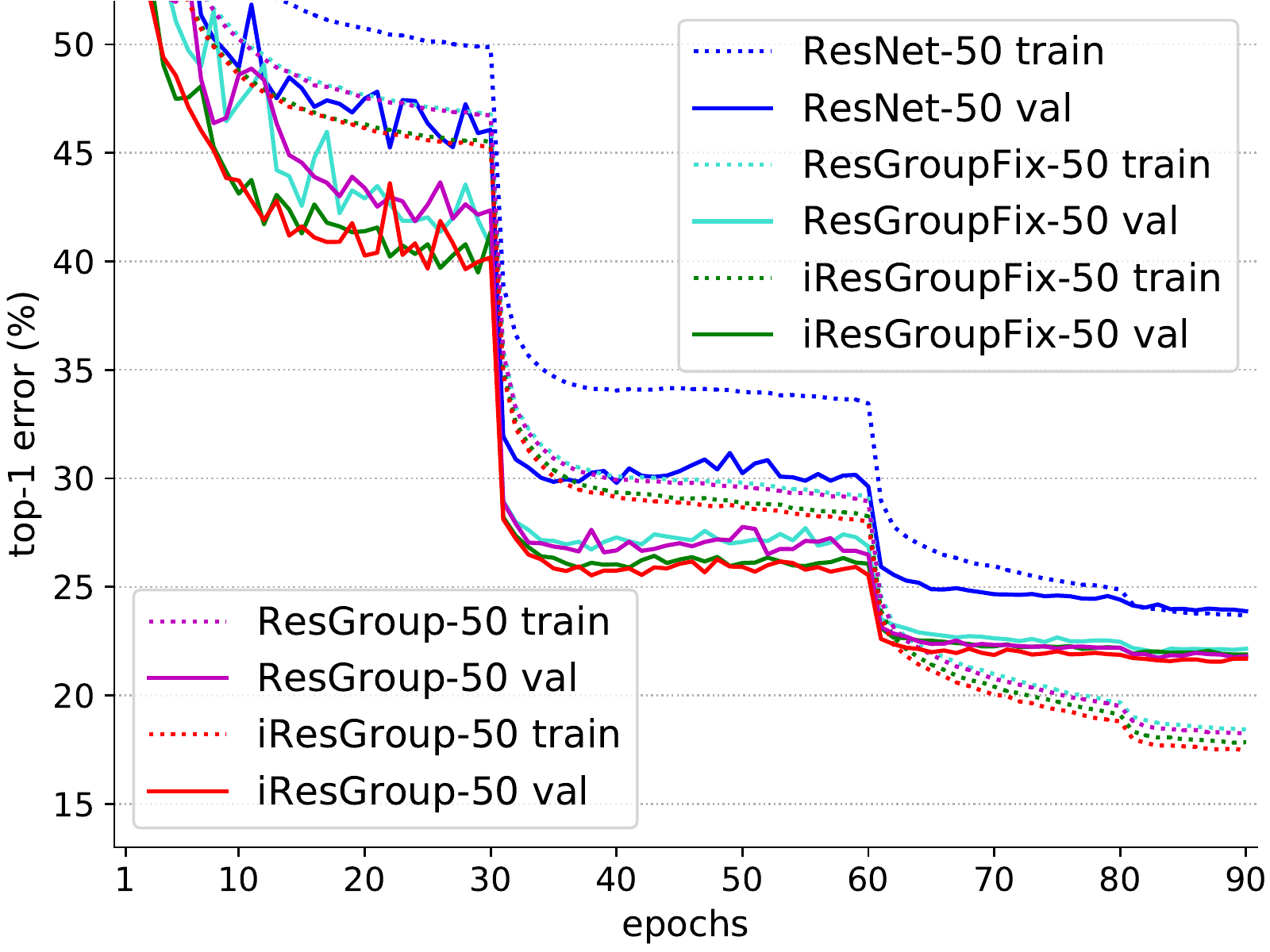}
\captionof{figure}{\small ResGroup curves.\label{fig:curvers_group}}
}
\end{table*}

\begin{table}[t]
\vspace{-0.3in}
\centering
\caption{Video recognition error rates (\%), parameters are in millions.}
\label{table:res_video}
\begin{tabular}{l|cccc|cccc}
\hline
\multirow{2}{*}{Network}& \multicolumn{4}{c|}{Kinetics-400} & \multicolumn{4}{c}{Something-Something-v2} \\ \cline{2-9}
&top-1&top-5&params&{\small GFLOPs}&top-1&top-5&params&{\small GFLOPs}\\ \hline
baseline3D-50 \cite{he2016deep}& 37.01  & 15.41&47.00&93.26 & 46.50 &19.02&  46.54&93.26 \\
iResNet3D-50&   \bf 33.91& \bf13.36&  47.00&93.93&\bf45.56  &\bf17.73 &  46.54&93.93 \\
\hline
\end{tabular}
\end{table}

\begin{table}[h!]
\centering
\caption{Classification error (\%) on CIFAR-10/100. For 164 layers train the model five times and show "best(mean$\pm$std)" as in \cite{he2016deep}.  P stands for parameters (in millions).}
\label{table:res_cifar-10}
\scalebox{0.9}{
\begin{tabular}{l|cc|cc|cc|cc}
\hline
\multirow{2}{*}{Network}&\multicolumn{2}{c|}{ 164 layers}&\multicolumn{2}{c|}{ 1001 layers}&\multicolumn{2}{c|}{ 2000 layers}&\multicolumn{2}{c}{ 3002 layers}\\ \cline{2-9}
&top-1&\scalebox{0.8}{P/GFLOPs}&top-1&\scalebox{0.8}{P/GFLOPs}&top-1&\scalebox{0.8}{P/GFLOPs}&top-1&\scalebox{0.8}{P/GFLOPs}\\ \hline
\hline
{\underline {CIFAR-10:}} & &&&&&&&\\
baseline \cite{he2016deep} &5.23 (5.54$\pm$0.37)& 1.70/0.26&7.43&10.33/1.59&fail&20.62/3.17&fail&30.93/4.75\\ 
iResNet &{\bf4.80} (5.00$\pm$0.14)&1.70/0.26 &\bf4.61&10.33/1.59 &\bf4.40&20.62/3.17 &\bf4.95&30.93/4.75\\ \hline
\hline
{\underline {CIFAR-100:}} & &&&&&&&\\
baseline \cite{he2016deep} &23.86 (24.48$\pm$0.39)& 1.73/0.26 & 26.98&10.35/1.59&fail&20.65/3.17&fail& 30.96/4.75\\ 
iResNet &{\bf22.26} (22.37$\pm$0.13)&1.73/0.26 &\bf20.92&10.35/1.59 &\bf21.12&20.65/3.17 &\bf21.46&30.96/4.75\\ \hline
\end{tabular}}
\end{table}

{\bf Results of iResNet on video recognition.} 
We perform comparison between our iResNet and the baseline ResNet on   large-scale video recognition, using the 50-layer network. Training a CNN on video recognition is even harder than in the case of images, as the number of parameters and complexity increase significantly. Our iResNet outperforms the baseline by a large margin, on both video datasets: Kinetics-400 and Someting-Something-v2 (see Table \ref{table:res_video}). For instance, on Kinetics-400, we report a 3.1\% improvement for top-1 error over the baseline. This shows that our approach helps the optimization process, and can give a greater benefit over the baseline with the increase of learning task difficulty. See the Appendix for training curves and network details.

{\bf Results of iResNet on CIFAR-10/100.} 
As these two datasets are much smaller than ImageNet and the networks have lower computational complexity, we can push further with the question regarding the depth limits and experiment with unprecedented depths for a CNN.  We report results on 164, 1001, 2000 and 3002 layers. iResNet outperforms the baseline for all depths. When increasing the depth from 164 to 1001, the baseline shows significant degradation issues, giving worse results. In contrast, our approach shows significant improvements. For depths 2000 and 3002, the baseline fails to converge, showing critical optimization issues.  In the case of iResNet, although the validation accuracy begins to decrease at a depth of  3002 (probably due to overfitting), the training loss shows no optimization issues, giving better results with increased depth (see the Appendix for training curves and more details). All of these show no degradation issues for iResNet.  Practically, these experiments may suggest that our approach is not bounded by increasing depth, the only limitations for learning increasingly powerful representations are stem from data and computational resources.

\addtolength{\tabcolsep}{-1pt}
\begin{table}[t]
\centering
\caption{Validation error rates (\%) comparison results of ResGroup on ImageNet.}
\label{table:resGroup_224}
\scalebox{0.93}{
\begin{tabular}{l|cccc|cccc|cccc}
\hline
\multirow{2}{*}{Network}&\multicolumn{4}{c|}{50 layers}&\multicolumn{4}{c|}{101 layers}&\multicolumn{4}{c}{152 layers}\\ \cline{2-13}
&top-1&top-5&\small params&\tiny GFLOPs&top-1&top-5&\small params&\tiny GFLOPs&top-1&top-5&\small params&\tiny GFLOPs\\ \hline
baseline \cite{he2016deep}&23.88&7.06&25.56&4.14&22.00&6.10&44.55&7.88 &21.55&5.74&60.19&11.62\\ 
ResNeXt \cite{xie2017aggregated}&  22.44 & 6.25  &  25.03 & 4.30  & 21.03  &  5.66 &  44.18 & 8.07   & 20.98  & 5.48  &  59.95  & 11.84 \\ 
ResGroupFix&  21.96 &  6.15 & 23.37  & 4.30  & 20.94  & 5.56  &  43.79&   8.33  & 20.70  &  5.48 &60.61   &12.35\\ 
ResGroup&  21.73 &  5.94 &   24.89  & 5.43   & 20.98  &  5.46 &  47.81  & 9.94     & 20.81  &  5.48 &   66.99  & 14.70 \\ 
iResGroupFix&   21.88& 5.99  &   23.37  & 4.47    &  20.92 &  5.54 &   43.79  & 8.49    & 20.75  &5.51   &   60.61  & 12.53\\ 
iResGroup&  \bf21.55 &   \bf5.75 &   24.89  & 5.60   &  \bf20.55  &    \bf5.45&   47.81  & 10.11   &   \bf20.34 & \bf5.20  &   66.99  & 14.87 \\ 
\hline
\end{tabular}}
\end{table}
\addtolength{\tabcolsep}{1pt}

{\bf Results of ResGroup on ImageNet.} Table \ref{table:resGroup_224} shows  results of ResGroup and ResGroupFix compared to baseline. Integrating the ideas from ResStage and the improved projection shortcut build improved ResGroup  and ResGroupFix (iResGroup and iResGroupFix). All proposed architectures significantly outperform  the baseline ResNet  \cite{he2016deep} and ResNeXt \cite{xie2017aggregated}. For instance, on depth 50, iResGroupFix yields a 2\% top-1 improvement over the baseline, while iResGroup improves the baseline by 2.33\%. Fig. \ref{fig:curvers_group} presents the training curves of our approaches compared to the baseline. 

\addtolength{\tabcolsep}{-1pt}
\begin{table*}[t]
\centering
\caption{Proposed backbones on SSD~\cite{liu2016ssd} object detector with $300$$\times$$300$ input image size (results on COCO val2017).}
\label{table:obj_det}
\scalebox{0.77}{
\begin{tabular}{l|ccc|ccc|ccc|ccc|cc}
\hline
\multirow{2}{*}{Backbone}& \multicolumn{3}{c|}{Avg. Precision, IoU:}& \multicolumn{3}{c|}{Avg. Precision, Area:}& \multicolumn{3}{c|}{Avg. Recall, \#Dets:}& \multicolumn{3}{c|}{Avg. Recall, Area:}&\multirow{2}{*}{\scalebox{0.8}{ params}}&\multirow{2}{*}{\scalebox{0.6}{ GFLOPs}} \\
\cline{2-13}
& 0.5:0.95 & 0.5&0.75&S&M&L&1&10&100&S&M&L&& \\
\hline
\hline
ResNet-50~\cite{he2016deep}& 26.20  &  43.97 &  26.96&   8.12&   28.22&   42.64&  24.50&   35.41&   37.07&   12.61&  40.76&     57.25&  22.89 & 20.92 \\
iResNet-50& 27.74  & 45.85  &28.51  & 8.52  & 30.07  & 44.62  &25.29  & 36.90  &38.51   & 13.28  &  42.79& 58.57    &   22.89&20.99  \\
iResGroupFix-50&  28.90 &   47.44&  29.99& 9.70  &  31.49 &  45.83 & 25.97 & 37.84  &  39.52 &14.63   &44.17  &  59.52   &  21.13 &18.61  \\
iResGroup-50&  \bf29.56 & \bf48.38  & \bf30.87 &  \bf10.33 & \bf32.52  &  \bf46.62 & \bf26.40 & \bf38.49  & \bf40.24  & \bf15.10  & \bf44.82 & \bf60.20    & 22.66  & 21.62 \\

\hline
   
ResNet-101~\cite{he2016deep}&29.58   & 47.69  & 30.80 &  9.38 &   31.96&   47.64&  26.47&   38.00&   39.64&   14.09& 43.54 &     61.03&   41.89& 48.45 \\
iResNet-101& 30.92  & 49.50  & 32.29 & 10.05  &  34.27 & 49.13  & 27.15 &  39.34 & 41.08  & 15.21  & 45.93 &  61.90   & 41.89  & 48.49 \\
iResGroupFix-101 &  31.64 &  50.70 & 33.28 & 11.21  &  34.91 &  50.20 & 27.94 &  40.41 &  42.22 & 16.84  & 46.99 &  63.64   & 41.55  &48.25  \\
iResGroup-101 & \bf32.81  & \bf51.78  & \bf34.55 &  \bf11.81 &  \bf36.56 & \bf51.72  & \bf28.37 & \bf41.43  &  \bf43.22 & \bf17.20  & \bf48.54 &  \bf64.08   &  45.58 & 54.87 \\
      
\hline
\end{tabular}}
\end{table*}
\addtolength{\tabcolsep}{1pt}

 {\bf Results on COCO object detection.} As we propose improved network architectures for general image classification, these networks can be used as backbones on other complex visual recognition tasks, such as object detection. We integrate our backbones in an object detection framework to show the contribution in recognition performance.  We use the single shot detector (SSD)~\cite{liu2016ssd} on a $300$$\times$$300$ input image size. As SSD is well known for efficiency and in \cite{liu2016ssd} the backbone  provides 1024 output feature maps, for maintaining a similar framework, we remove all the layers after the main stage 3 of our backbones and we set the stride to 1 in stage 3. Table~\ref{table:obj_det} shows that using our networks (iResNet, iResGroupFix and iResGroup) as backbones on SSD, improves significantly the results in comparison with the baseline ResNet~\cite{he2016deep} on all metrics. Besides, our networks maintain the efficiency. Notably, our iResGroup with 50 layers nearly matches in therms of recognition performance the ResNet baseline with 101 layers. These results show the importance for many other tasks of having powerful backbones.

 \addtolength{\tabcolsep}{+10pt}
 \begin{table}[t]
\centering
\caption{Single-crop error rates (\%) comparison with other networks  on ImageNet validation set. $^\dag$ some approaches use larger image crops than 320$\times$320, Inception family uses 299$\times$299. }
\label{table:state}
\scalebox{1}{
\begin{tabular}{l|cc|cc}
\hline
\multirow{2}{*}{Method}& \multicolumn{2}{c|}{224$\times$224}&\multicolumn{2}{c}{320$\times$320$^\dag$} \\ \cline{2-5}
&top-1&top-5&top-1&top-5\\ \hline
ResNet-200 \cite{he2016identity}& 21.7 &5.8& 20.1 &4.8 \\
Inception-v3 \cite{szegedy2016rethinking}& -& -& 21.2 &5.6\\
Inception-v4 \cite{szegedy2017inception} &- &- &20.0& 5.0\\
Inception-ResNet\cite{szegedy2017inception} &-& -& 19.9& 4.9\\
DenseNet-264 \cite{huang2017densely} &22.15 &6.12 &-& -\\
Attention-92 \cite{wang2017residual}& -& - &19.5& 4.8\\
NASNet-A \cite{zoph2018learning}& -& -& 17.3 &3.8 \\
SENet-154 \cite{hu2018squeeze}&18.68& 4.47 &17.28 &3.79\\
\hline
iResNet-200&20.52&5.36&19.36&4.56 \\
iResNet-404& 20.30&  5.26&19.35& 4.61 \\
iResGroup-152&20.34&5.20&19.09&4.59\\
\end{tabular}}
\end{table}
\addtolength{\tabcolsep}{-10pt}
{\bf Comparison to other networks on ImageNet.} Table \ref{table:state} shows that our approach is competitive to other powerful network architectures on ImageNet. There are some approaches, for instance \cite{zoph2018learning,hu2018squeeze} with better results. However, these approaches use longer training  schedules, a bigger training crop \cite{zoph2018learning} and/or more complex network building blocks. For instance, \cite{hu2018squeeze} uses a squeeze-and-excitation block to improve the results, which increases the model complexity. Importantly,  many of these works are based on ResNet baseline, therefore, they will directly benefit from our proposed improvements. We leave this exploration for future work.

\section{Conclusion}
This work proposed an improved version of residual networks. Our improvements address all three main components of a ResNet: information propagation through the network, the projection shortcut, and the building block. We report consistent improvements over the baseline. For instance, on the widely used 50 layers deep ResNet, we present improvements in top-1 error, in different settings, ranging from 1.19\% to 2.33\%. These improvements are obtained without increasing model complexity. Our proposed approach facilitates learning of extremely deep networks, showing no optimization issues when training networks with over 400 layers (on ImageNet) and over 3000 layers (on CIFAR-10/100).

\appendix
\section{Appendix\label{sec:appendix}}

In this Appendix we present additional explanations and experimental setup details. As we pointed out, on all six datasets we follow the common training and testing procedures established by the research community.  All our models are trained from scratch (except on object detection, where we pre-train on ImageNet), using the weights initialization of \cite{he2015delving} for all convolutional layers.

\subsection{iResNet on ImageNet}
The number of building blocks of ResNet-based architectures depends on the number of layers. Table~\ref{table:block} presents the number of bottleneck blocks for each stage of the network, for the 50-, 101-, 152-, 200-, 302-, 404-layers deep networks. 
\addtolength{\tabcolsep}{2.5pt}
\begin{table}[h]
\centering
\caption{Bottleneck building blocks per stage for each considered depth on ImageNet.}
\label{table:block}
\begin{tabular}{l|cccccc}
\hline
& 50 layers& 101 layers& 152 layers& 200 layers& 302 layers& 404 layers \\ \hline
Stage 1& 3 &3&3&3&4&4 \\
Stage 2& 4& 4&8&24&34&46 \\
Stage 3&6&23&36&36&58&80\\
Stage 4&3&3&3&3&4&4\\
\hline
\end{tabular}
\end{table}
\addtolength{\tabcolsep}{-2.5pt}

\begin{figure*}[t]
\centering
\begin{tabular}{cccc}
\subfloat{\includegraphics[width=0.26\textwidth]{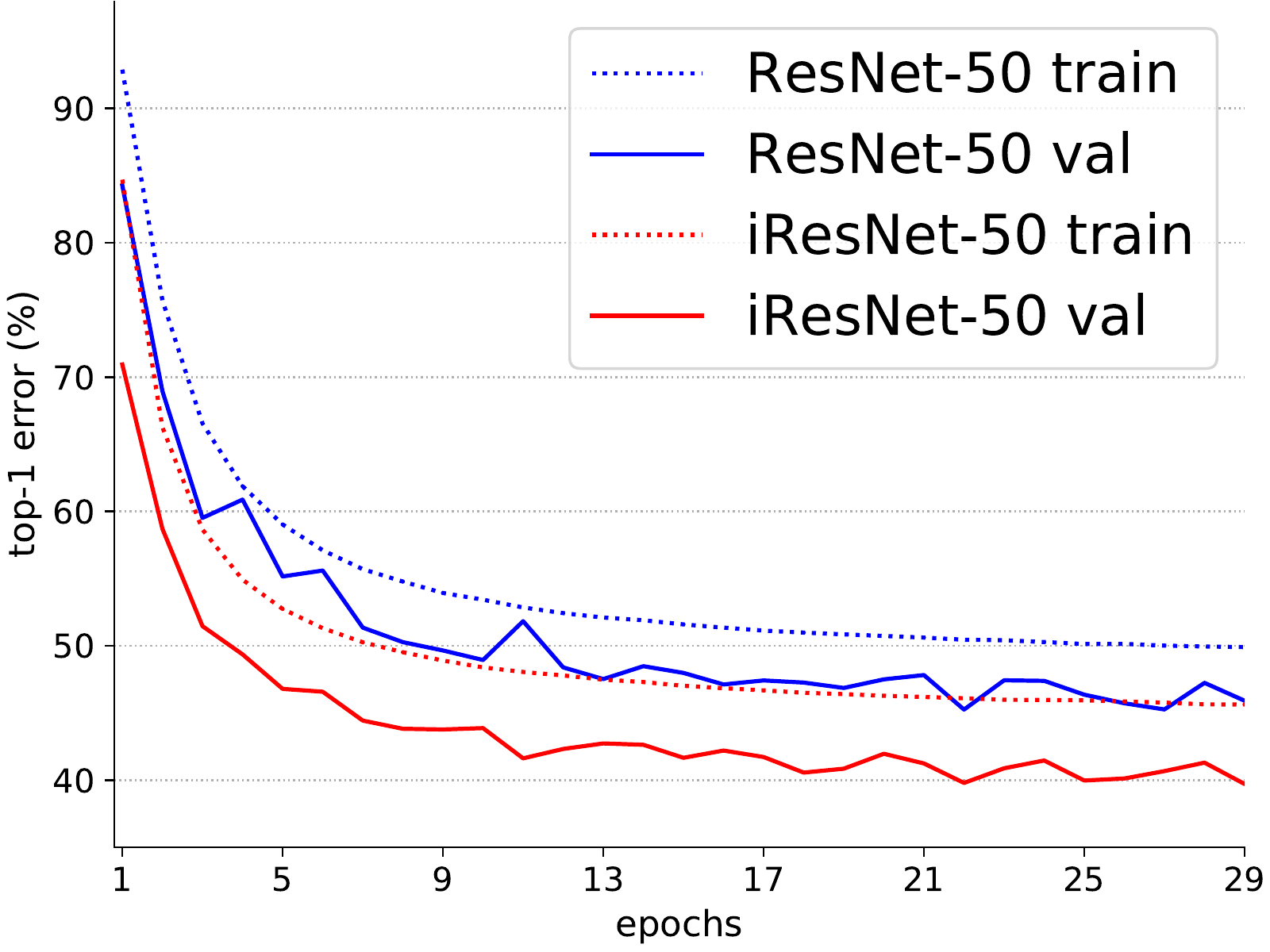}} &
\subfloat{\includegraphics[width=0.26\textwidth]{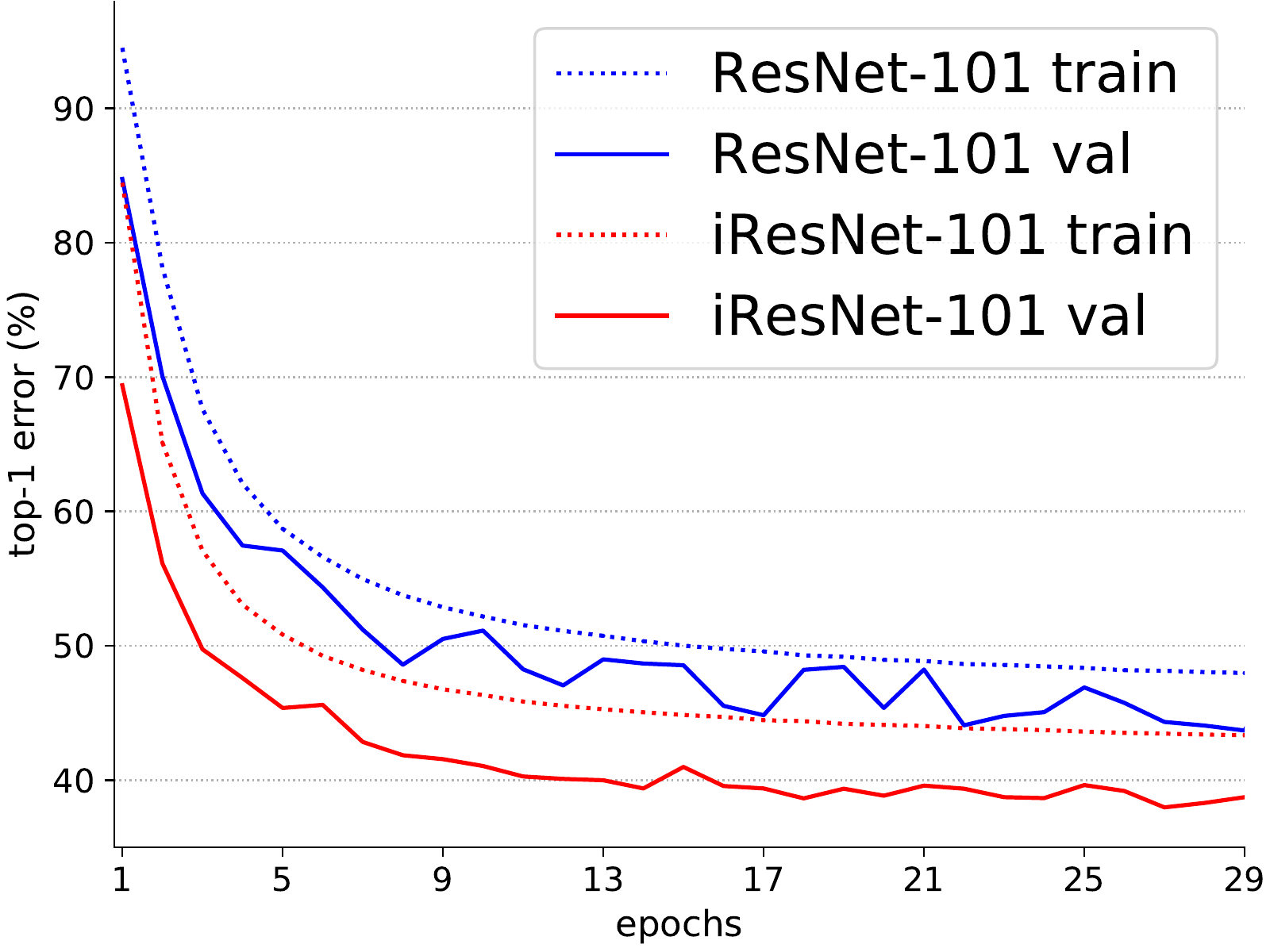}} &
\subfloat{\includegraphics[width=0.26\textwidth]{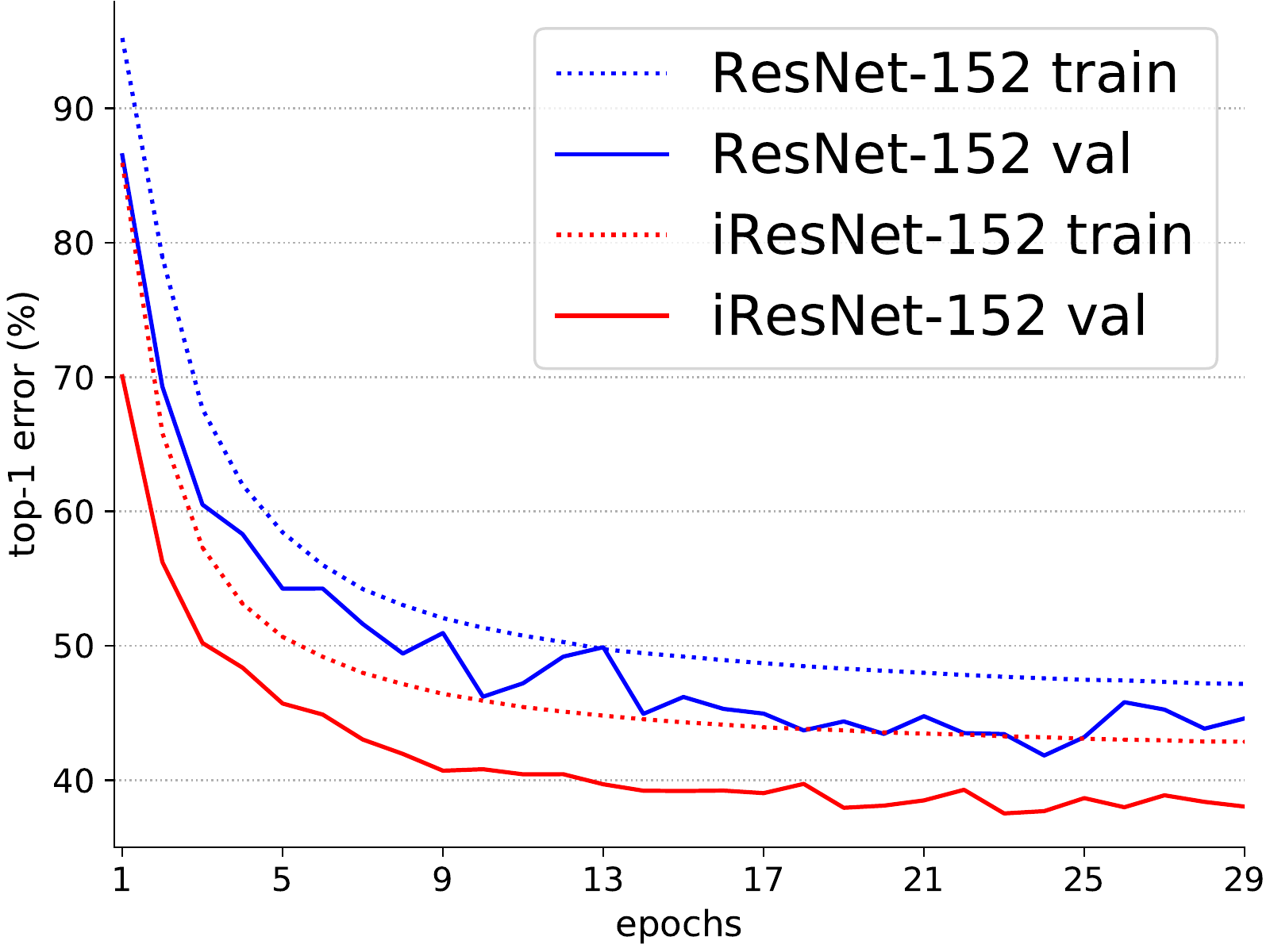}} &
\subfloat{\includegraphics[width=0.26\textwidth]{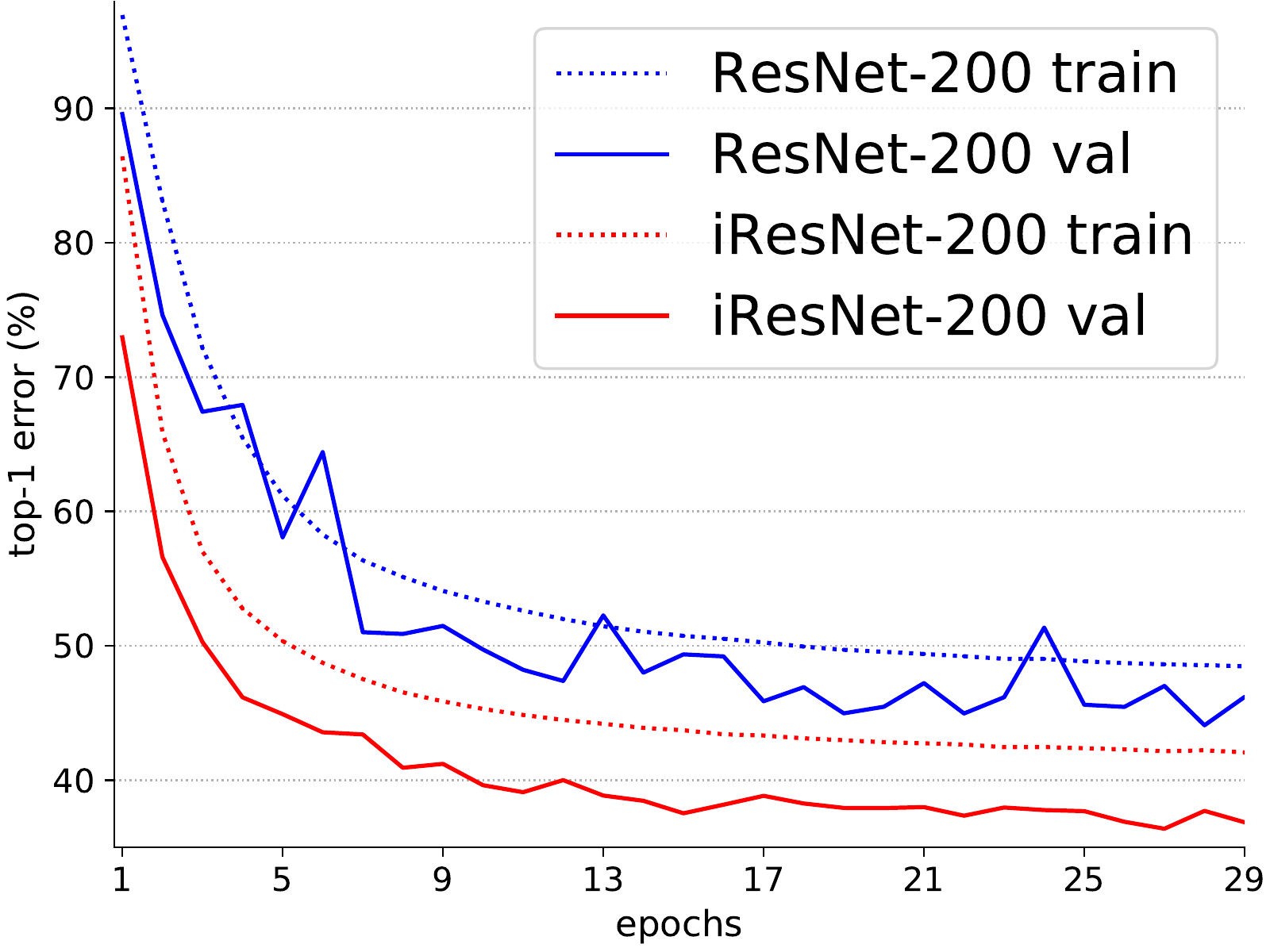}}
\end{tabular}
\caption{The beginning of training and validation curves on ImageNet for ResNet and iResNet on 50, 101, 152 and 200 layers.}
\label{fig:curves_iresnet_appendix}
\end{figure*}

In the main paper, we present the results with a focus on the final training and validation performance. However, something very suggestive occurs at the beginning of training. We can see from Fig.~\ref{fig:curves_iresnet_appendix} that the original ResNets \cite{he2016deep} converge  very slowly at the beginning of training. With increasing depth, the difficulty of convergence increases as well. This can easily be seen from the training and validation curves, where the training performance degrades with the increasing  number of layers.  In contrast, for our iResNets the starting convergence is not affected by increasing the depth. From the training curves, we can see the iResNets learn much faster than the original ResNets.  Besides, we can also notice from the graphs that, during training, the validation accuracy of iResNets has significantly lower random fluctuations than the baseline, with the curves descending in a smoother way, suggesting that the learning is also more stable. This aspect is visible in Fig. \ref{fig:curves_iresnet_appendix}, and also in  Fig. 4 of the main paper, where we show different depths of the networks alongside each other. 

\addtolength{\tabcolsep}{-4pt}
\addtolength{\tabcolsep}{6pt}

\begin{table}[h]
\centering
\caption{Error rates (\%) for 320$\times$320 validation crop size on ImageNet}
\label{table:res_320}
\begin{tabular}{l|cccc|cccc}
\hline
&\multicolumn{4}{c|}{depth 50}&\multicolumn{4}{c}{depth 101}\\ \cline{2-9}
&top-1&top-5&params&\small GFLOPs&top-1&top-5&params&\small GFLOPs\\ \hline
baseline \cite{he2016deep}&22.69&6.24&25.56&8.45& 20.85&5.28 &44.55&16.07\\ 
pre-activation \cite{he2016identity}&22.67&6.22&25.56&8.45& 20.83&5.45&44.55&16.07\\ 
ResStage&22.07&5.90&25.56&8.45&20.46&5.19& 44.55&16.07\\ 
iResNet &\bf21.35&\bf5.70&25.56&8.53& \bf20.12&\bf4.97&44.55&16.15\\ 
\hline
\hline
&\multicolumn{4}{c|}{depth 152}&\multicolumn{4}{c}{depth 200}\\ \cline{2-9}
&top-1&top-5&params&\small GFLOPs&top-1&top-5&params&\small GFLOPs\\ \hline
baseline \cite{he2016deep}&20.35&5.05&60.19&23.71&21.24&5.52&64.67&30.93\\ 
pre-activation \cite{he2016identity}&20.32&5.06&60.19&23.71&20.30&5.03&64.67&30.93\\ 
ResStage &19.81&4.73&60.19&23.71&19.66&4.62&64.67&30.93\\  
iResNet &\bf19.59&\bf4.62&60.19&23.78&\bf19.36&\bf4.56&64.67&30.99\\  \hline
\end{tabular}
\end{table}

Another important observation that we can make based on the training curves (see, for instance, Fig. 4 in the main paper) is that iResNets show consistent and significant improvements on the validation results. Over and above this, the results on the training set convergence for iResNets show even more significant benefits over the baseline ResNets \cite{he2016deep}, indicating the efficiency of optimization. This may suggest that using additional regularization techniques, such as dropout \cite{hinton2012improving}, may result in even greater improvements for our proposed approaches. We leave this exploration for future work.

While we train all models on ImageNet with the standard procedure of 224$\times$224 (and use the same crop size for validation) as in \cite{he2016deep,he2016identity}, performing only the validation on a 320$\times$320 crop size, as in \cite{he2016identity}, can improve the results considerably. This is easily possible because of the nature of ResNets, which are fully convolutional.  Table \ref{table:res_320} presents the results with the 320$\times$320 validation crop.

\addtolength{\tabcolsep}{-3pt}
\begin{figure}[t]
\centering
\begin{tabular}{cc}
\subfloat{\includegraphics[width=0.50\textwidth]{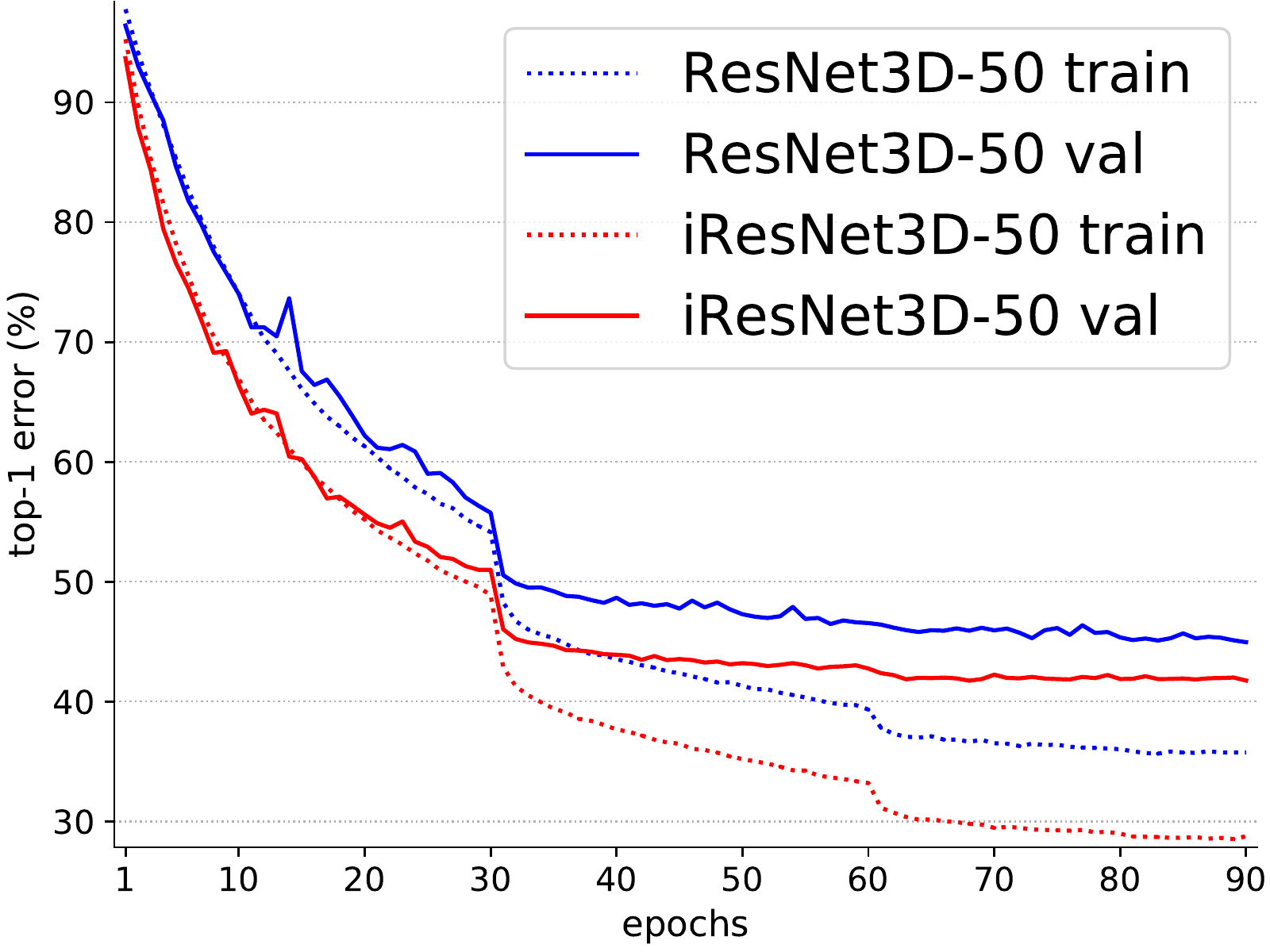}} &
\subfloat{\includegraphics[width=0.50\textwidth]{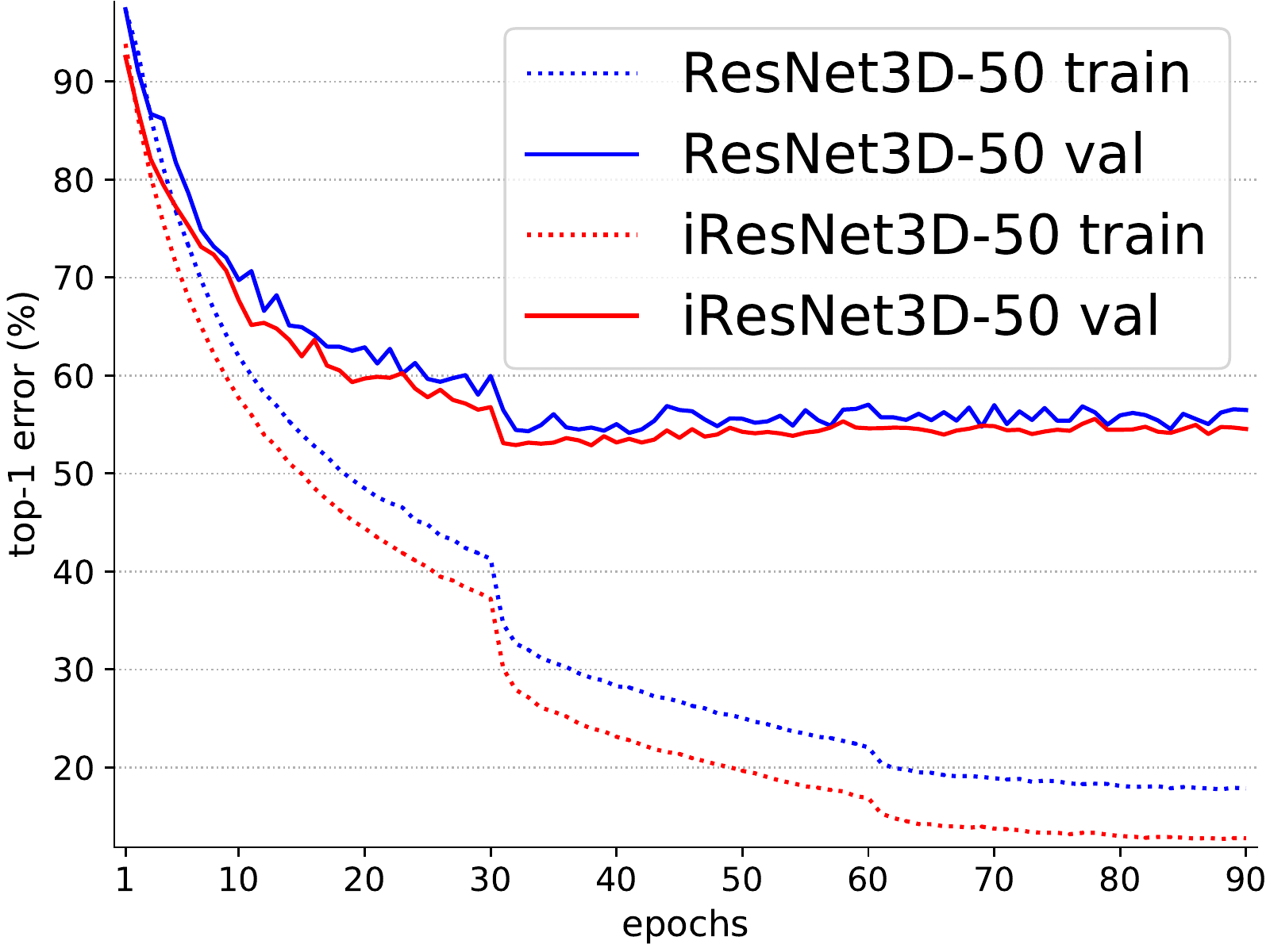}}
\end{tabular}
\caption{Training curves on Kinetics-400 (left) and Something-Something-v2 (right). These are the results computed during training over independent clips.}
\label{fig:video_curves}
\end{figure}
\addtolength{\tabcolsep}{3pt}

\begin{table}[h!]
\centering
\caption{ResNet3D architecture for video recognition.}
\label{table:net_video}
\begin{tabular}{c|c|c}
\hline
stage                  & output                 & ResNet3D-50                       \\ \hline
\multirow{2}{*}{starting}                 & 16$\times$112$\times$112                & 5$\times$7$\times$7, 64, stride (1,2,2)     \\  \cline{2-3}
 & 16$\times$56$\times$56 & 1$\times$3$\times$3 max pool, stride (1,2,2) \\  \hline
1&16$\times$56$\times$56 & \bsplitcell{1$\times$1$\times$1, 64\\ 3$\times$3$\times$3, 64\\ 1$\times$1$\times$1, 256}$\times$3  \\ \hline

2 & 16$\times$28$\times$28 & \bsplitcell{1$\times$1$\times$1, 128\\ 3$\times$3$\times$3, 128\\ 1$\times$1$\times$1, 512}$\times$4 \\ \hline

3 & 8$\times$14$\times$14 & \bsplitcell{1$\times$1$\times$1, 256\\ 3$\times$3$\times$3, 256\\ 1$\times$1$\times$1, 1024}$\times$6  \\ \hline

4 & 4$\times$7$\times$7 & \bsplitcell{1$\times$1$\times$1, 512\\ 3$\times$3$\times$3, 512\\ 1$\times$1$\times$1, 2048}$\times$3  \\ \hline
ending & 1$\times$1$\times$1 &  \begin{tabular}[c]{@{}c@{}}global avg pool\\ 400/174-d fc\end{tabular}  \\ \hline
\end{tabular}
\end{table}

\subsection{iResNet on video recognition}

The training and validation curves for the Kinetics-400 and Something-Something-v2 datasets are visualized in Fig. \ref{fig:video_curves}. Note that the results of these curves are obtained by computing the top-1 error independently for each video clip. From the curves, we can see that iResNet3D facilitates learning.

The ResNet network architecture used for video recognition  is presented in Table \ref{table:net_video}. We use 16-frame input clips.  For training,  we randomly select the input clips from the video. For Kinetics-400, we also skip four frames to cover a longer video period within a clip. As the  Something-Something-v2 dataset contains much shorter videos, we skip only one frame. The spatial size is 224$\times$224, randomly cropped from a scaled video, where the shorter side is randomly selected from the interval [256, 320], similar to \cite{simonyan2014very,wang2018non}. Different from image datasets, where we do not use any other regularization besides weight decay, as the networks on video data are prone to overfitting due to the increase in number of parameters, on the video datasets we use dropout \cite{hinton2012improving} after the global average pooling layer, with a 0.5 dropout ratio. For the final validation, following common practice, we uniformly select a maximum of 10 clips per video. Each clip is scaled  to 256 pixels for the shorter spatial side. We take 3 spatial crops to cover the spatial dimensions. In total, this results in a  maximum  of 30 clips per video, for each of which we obtain a prediction, we average the softmax scores to get the final prediction for a video.

\addtolength{\tabcolsep}{-2pt}
\begin{figure*}[t]
\centering
\begin{tabular}{cc}
\subfloat{\includegraphics[width=0.5\textwidth]{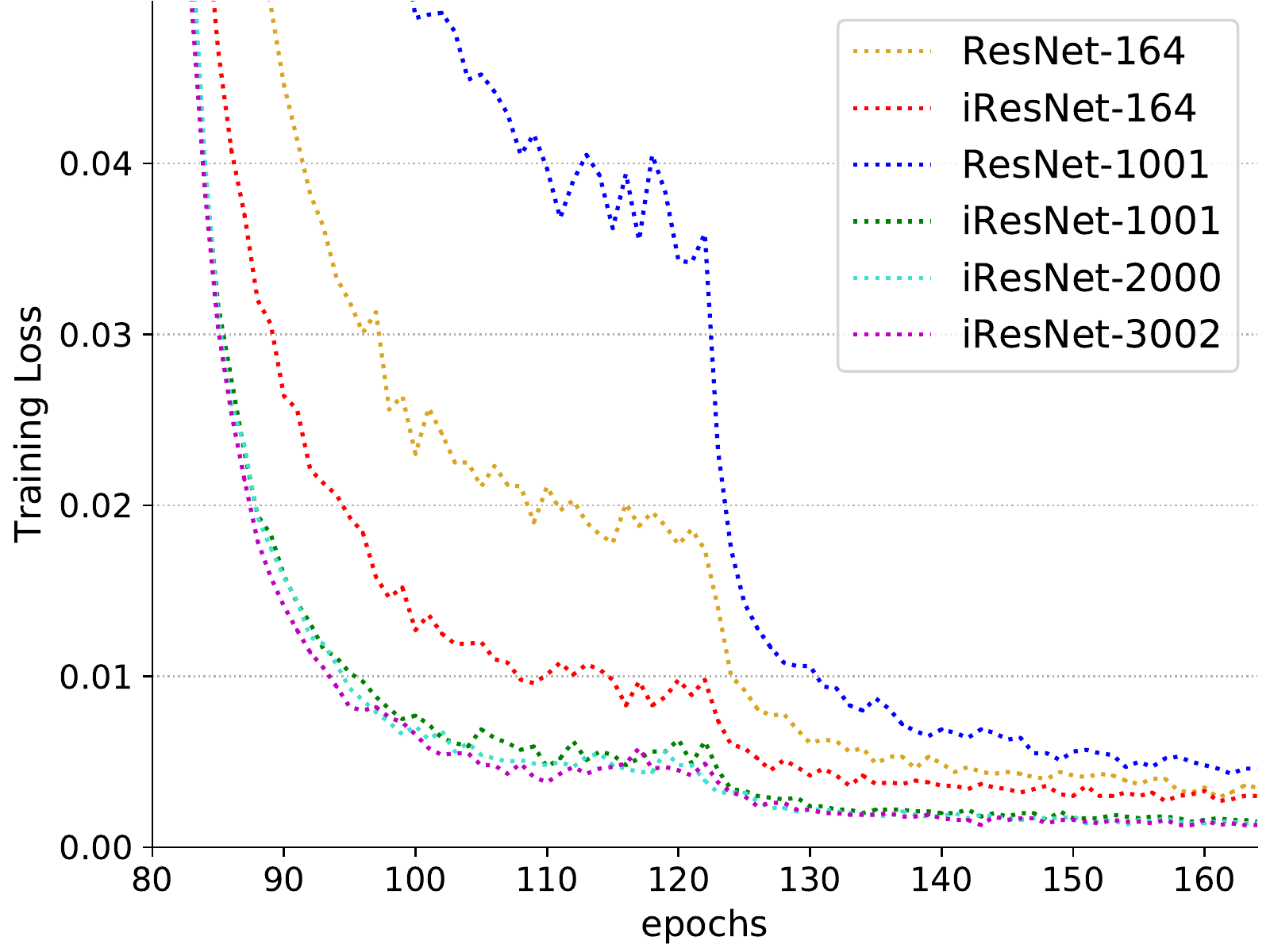}} &
\subfloat{\includegraphics[width=0.5\textwidth]{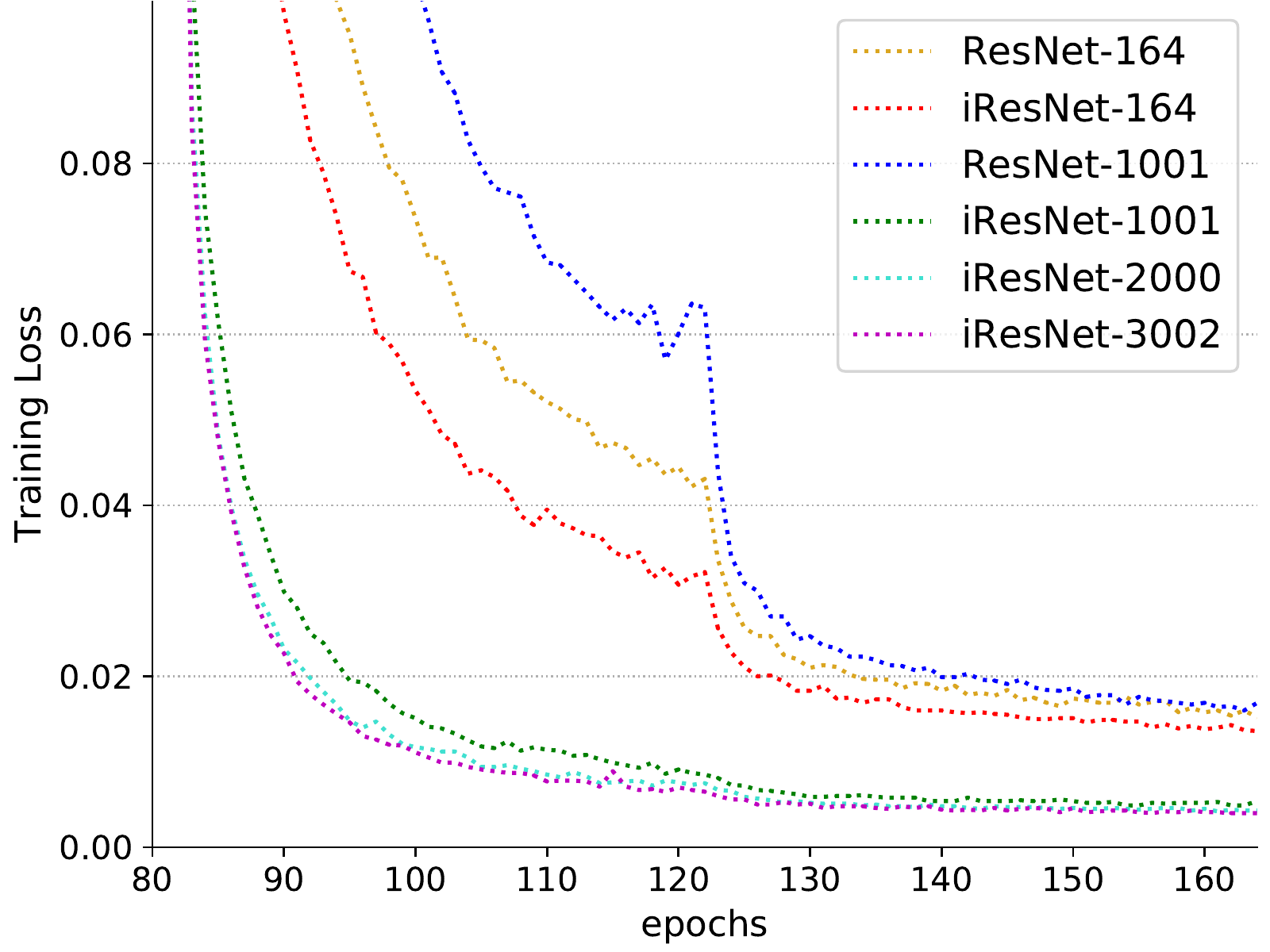}}
\end{tabular}
\caption{Training Loss on CIFAR-10 (left), CIFAR-100 (right).}
\label{fig:cifar}
\end{figure*}
\addtolength{\tabcolsep}{7pt}

\subsection{iResNet on CIFAR-10/100}
The degradation problem is best visible on the training loss, as on the validation, the performance may be influenced not only by optimization issues but also by overfitting, for which there are other techniques for addressing it.
Very suggestive are the training loss curves on CIFAR-10/100, illustrated in Fig. \ref{fig:cifar}. We can see that, for the original ResNet \cite{he2016deep}, the degradation problem starts to be easily visible when increasing the depth from 164 to 1001, resulting in a worse training loss. For deeper networks, such as 2000 or 3002, the  original ResNet faces critical optimization issues, not even starting to converge. In contrast, the iResNet training loss improves continuously from a depth of 164 to 3002, showing the success of optimization. We can also see less large fluctuations during training for our iResNets in comparison to baseline ResNets. .

%
\begin{table}[t]
\centering
\caption{ResNet architecture for CIFAR-10/100}
\label{table:net_cifar}
\begin{tabular}{c|c|c}
\hline
stage                  & output                 & ResNet                  \\ \hline
starting                 & 32$\times$32                & 3$\times$3, 16, stride 1    \\  \cline{1-3}
1&32$\times$32 & \bsplitcell{1$\times$1, 16\\ 3$\times$3, 16\\ 1$\times$1, 64}$\times n_1$  \\ \hline

2 & 16$\times$16 & \bsplitcell{1$\times$1, 32\\ 3$\times$3, 32\\ 1$\times$1, 128}$\times n_2$ \\ \hline

3 & 8$\times$8 & \bsplitcell{1$\times$1, 64\\ 3$\times$3, 64\\ 1$\times$1, 256}$\times n_3$  \\ \hline

ending & 1$\times$1 &  \begin{tabular}[c]{@{}c@{}}global avg pool\\ 10/100-d fc\end{tabular}  \\ \hline
\end{tabular}
\end{table}

\begin{table}[t]
\centering
\caption{Bottleneck building blocks per stage for each considered depth on CIFAR-10/100.}
\label{table:block_cifar}
\begin{tabular}{l|cccc}
\hline
& 164 layers& 1001 layers& 2000 layers& 3002 layers \\ \hline
Stage 1 ($\times n_1$ )&18& 111 &222&333 \\
Stage 2 ($\times n_2$ )&18& 111& 222&334 \\
Stage 3 ($\times n_3$ )&18&111&222&333\\
\hline
\end{tabular}
\end{table}

The networks for  CIFAR-10/100 are constructed as in \cite{he2016deep} (see Table \ref{table:net_cifar}). The training is performed following~\cite{he2016deep};  164 epochs, with a learning rate of 0.1, reduced by 1/10 at the  81-st and 122-nd epochs. However, we use a mini-batch size of 64 and only one GPU for training each model. As the image resolution on CIFAR-10/100 is 32$\times$32, the networks contain only 3 main stages (see Table~\ref{table:net_cifar}).  The number of building blocks for each main stage is presented in Table \ref{table:block_cifar} for the considered depths in this work (164-, 1001-, 2000-, 3002-layers).

\subsection{Comparison on the projection shortcut}

\begin{table}[t]
\centering
\caption{Projection shortcut comparison on ImageNet.}
\label{table:comp_proj}
\begin{tabular}{l|cccc}
\hline
&top-1(\%)&top-5(\%)&params&{\small GFLOPs}\\ \hline
ResNet~\cite{he2016deep}&23.88&7.06&25.56&4.14 \\
ResNet~\cite{he2019bag}& 23.26& 6.80& 25.56&4.14 \\
ResMax (ours) &\bf 22.85&\bf 6.42&25.56&4.18\\
\hline
\end{tabular}
\end{table}

As we pointed in the related work section of the main paper, although we present different directions of improvement than the collected refinements in~\cite{he2019bag}, there is an overlap with one point of our proposed contributions, regarding the projection shortcut.  
There are three main differences between our proposed projection shortcut and the one used in~\cite{he2019bag}. (1) We use max pooling (instead of average pooling as in~\cite{he2019bag}), this helps on improving the translation invariance of the network and ultimately improves the recognition performance. Also, max pooling is more suitable for performing local pooling of information (by taking the highest activation), while the average pooling is performing well for gathering global information  (as on larger regions there can be many high activations, and an average of them may describe better the information).
(2) We use a kernel of 3x3 for pooling (while~\cite{he2019bag} uses 2x2). In this way  the size of our kernel for max pooling is equal to the size of the spatial convolution ($3$$\times$$3$ conv) from the "Start Block". This is important, as it ensures that the element-wise addition (between the projection shortcut and the main block) is performed between the elements computed over the same spatial window (this is motivated also in the main paper, see Section 3.2). 
(3)  Different from ~\cite{he2019bag}, we integrate the proposed projection shortcut also on the first stage (while  ~\cite{he2019bag} uses their projection shortcut only on the last three stages). Since we proposed to use a  projection shortcut with 3x3 max pooling on the "Start Block" of each stage, we do not need to keep the original max pooling of the ResNet. Different from~\cite{he2019bag}, we also provide additional motivations for our proposed projection shortcut (see Section 3.2 for details). Table~\ref{table:comp_proj} presents the comparison results on the projection shortcut with~\cite{he2019bag} using a 50-layer deep network. Our approach outperforms~\cite{he2019bag} by a large margin. We call our network ResMax,  when using only our proposed projection shortcut on the ResNet baseline. 

\subsection{Discussion}
Someone may say: "Your work improved convolutional neural networks, that's good and useful in practice for reasonable depths such as 50 and 101. But what about these outrageous depths: over 400 or over 3000 layers!? What is the meaning of this and how are these unrealistic depths going to  be useful in practice?" As we pointed, depth is a key factor for learning powerful representations, and what is unpractical today, tomorrow may be ordinary and irreplaceable. Human brain contains a tremendous number of neurons, connections between them and layers, unmatched by any artificially built learning system. And yet, the human brain is able to process information extremely quickly, using very low energy, in comparison with our artificially built systems where the power consumption is significantly higher and falls far behind human level intelligence as performance. The human brain is able to achieve this impressive performance by not using its full capacity at the same time.  One possible scenario that we see in the not-so-distant future is the following: we will be able to build neural networks very close to the capacity of the human brain, but these unrealistically (today) huge networks will not be static. The neural network can dynamically change and  choose at the run-time which neurons, connections and layers  to use, based, for instance, on the complexity of the input data, task to solve and/or previous activations. This will tremendously reduce the computational complexity and power consumption. But the learning of deep and powerful representations may have "no limits".  Having systems that can easily learn without being negatively affected by  significantly increasing depth can play a critical factor in achieving the ultimate goal, human level intelligence.

\clearpage
%
%
\bibliographystyle{splncs04}
\bibliography{refs_short}
\end{document}